\documentclass{article}
\def\NeurIPS{}
\usepackage[utf8]{inputenc} 
\usepackage[T1]{fontenc}    
\usepackage{booktabs}       
\usepackage{nicefrac}         
\usepackage{microtype}      
\usepackage[dvipsnames]{xcolor}         
\usepackage[round]{natbib}

\usepackage{amssymb,amsmath,amsthm,bbm}
\usepackage{verbatim,float,url,dsfont}
\usepackage{algorithm,algorithmic}
\usepackage{mathtools,enumitem}
\usepackage{multirow,multicol}
\usepackage{ragged2e}
\usepackage{xr-hyper}
\usepackage{array}
\usepackage{stmaryrd,scalerel}

\ifdefined\NeurIPS  
\usepackage{hyperref}       
\else 
\usepackage[colorlinks=true,citecolor=blue,urlcolor=blue,linkcolor=blue]{hyperref}
\usepackage[margin=1in]{geometry}
\fi

\usepackage{threeparttable}  
\usepackage{subcaption} 
\usepackage{scrextend} 
\usepackage{arydshln} 

\newtheorem*{assumption*}{\assumptionnumber}
\providecommand{\assumptionnumber}{}


\def\R{\mathbb{R}}

\def\P{\mathbb{P}}

\def\th{^{\mathrm{th}}}

\def\cC{\mathcal{C}}

\def\cI{\mathcal{I}}
\def\cJ{\mathcal{J}}

\def\cT{\mathcal{T}}

\def\cX{\mathcal{X}}
\def\cY{\mathcal{Y}}

\def\ind#1{\mathds{1}\left\{#1\right\}}

\usepackage{chngcntr}
\usepackage{apptools}
\AtAppendix{\counterwithin{theorem}{section}}
\AtAppendix{\counterwithin{proposition}{section}}

\usepackage{mathtools}

\usepackage{mdframed}
\definecolor{light-gray}{HTML}{F7F7F7}
\definecolor{frame-color}{HTML}{CFCFCF}
\surroundwithmdframed[
backgroundcolor=light-gray,%
linecolor=frame-color%
]{minted}

\usepackage{mathtools}
\mathtoolsset{showonlyrefs}

\newtheorem{proposition}{Proposition}
\theoremstyle{definition}

\def\qhat{\hat{q}} 
\def\qtilde{\tilde{q}} 
\def\fhat{f}
\def\hhat{\hat{h}} 

\def\navg{n_{\mathrm{avg}}}
\def\Xtest{X_{\mathrm{test}}}
\def\Ytest{Y_{\mathrm{test}}}
\def\nullclust{\mathsf{null}}
\def\softmax{\mathsf{softmax}}
\def\APS{\mathsf{APS}}
\def\RAPS{\mathsf{RAPS}}
\def\standard{\textsc{standard}}
\def\classwise{\textsc{classwise}}
\def\clustered{\textsc{clustered}}

\newcommand{\quantile}{\mathrm{Quantile}}

\newcommand{\attn}[1]{\textbf{#1}}

    \usepackage[final]{neurips_2023}

\title{Class-Conditional Conformal Prediction \\ with Many Classes} 

\author{%
  Tiffany Ding \\
  University of California, Berkeley \\
  \texttt{tiffany\_ding@berkeley.edu} \\
  \And
  Anastasios N.\ Angelopoulos \\
  University of California, Berkeley \\
  \texttt{angelopoulos@berkeley.edu} \\
  \And
  Stephen Bates \\
  MIT \\
  \texttt{s\_bates@mit.edu} \\
  \And
  Michael I.\ Jordan \\
  University of California, Berkeley \\
  \texttt{jordan@cs.berkeley.edu} \\
  \And
  Ryan J.\ Tibshirani \\
  University of California, Berkeley \\
  \texttt{ryantibs@berkeley.edu} \\
}

\begin{document}
\maketitle

\begin{abstract}
Standard conformal prediction methods provide a marginal coverage guarantee,
which means that for a random test point, the conformal prediction set contains 
the true label with a user-specified probability. In many classification
problems, we would like to obtain a stronger guarantee---that for test points
\emph{of a specific class}, the prediction set contains the true label with the
same user-chosen probability. For the latter goal, existing conformal prediction
methods do not work well when there is a limited amount of labeled data per
class, as is often the case in real applications where the number of classes is
large. We propose a method called \emph{clustered conformal prediction} that
clusters together classes having ``similar'' conformal scores and performs 
conformal prediction at the cluster level. Based on empirical evaluation across
four image data sets with many (up to 1000) classes, we find that clustered
conformal typically outperforms existing methods in terms of class-conditional
coverage and set size metrics.
\end{abstract}

\section{Introduction}
Consider a situation in which a doctor relies on a machine learning system that
has been trained to output a set of likely medical diagnoses based on CT
images of the head. Suppose that the system performs well on average and is
able to produce prediction sets that contain the actual diagnosis with at least 0.9 probability. Upon closer examination, however, the doctor discovers
the algorithm only predicts sets containing common and relatively benign
conditions, such as $\{\mathsf{normal}, \mathsf{concussion}\}$, and the sets
never include less common but potentially fatal diseases, such as
$\mathsf{intracranial\ hemorrhage}$. In this case, despite its high marginal
(average) performance, the doctor would not want to use such an algorithm
because it may lead to patients missing out on receiving critical care.  The
core problem is that even though the average performance of the algorithm is
good, the performance for some classes is quite poor.

Conformal prediction \citep{vovk2005algorithmic} is a method for producing 
set-valued predictions that serves as a wrapper around existing prediction
systems, such as neural networks. Standard conformal prediction takes a
black-box prediction model, a calibration data set, and a new test example 
$\Xtest \in \cX$ with unknown label $\Ytest \in \cY$ and creates a prediction
set $\cC(\Xtest) \subseteq \cY$ that satisfies \emph{marginal coverage}: 
\begin{equation} 
\label{eq:marginal-coverage}
\P(\Ytest \in \cC(\Xtest)) \geq 1- \alpha,
\end{equation}
for a coverage level $\alpha \in [0,1]$ chosen by the user.  However, as the
example above shows, the utility of these prediction sets can be limited in some
real applications. In classification, which we study in this paper, the label
space $\cY$ is discrete, and it is often desirable to have
\emph{class-conditional coverage}:            
\begin{equation}
\label{eq:class-conditional-coverage}
\P(\Ytest \in \cC(\Xtest) \mid \Ytest = y) \geq 1- \alpha, \quad \text{for all }
y \in \cY,  
\end{equation}
meaning that every class $y$ has at least $1-\alpha$ probability of being
included in the prediction set when it is the true label. Note that  
\eqref{eq:class-conditional-coverage} implies \eqref{eq:marginal-coverage}. 
Predictions sets that only satisfy \eqref{eq:marginal-coverage} may neglect the
coverage of some classes, whereas the predictions sets in
\eqref{eq:class-conditional-coverage} are effectively  ``fair'' with respect to
all classes, even the less common ones.  

Standard conformal prediction, which we will refer to as $\standard$, does not
generally provide class-conditional coverage. We present a brief case study to
illustrate.  

\paragraph{ImageNet case study.}

Running $\standard$ using a nominal coverage level of 90\% on 50,000 examples  
sampled randomly from ImageNet \citep{russakovsky2015imagenet}, a large-scale
image data set described later in Section \ref{sec:experiments}, yields
prediction sets that achieve very close to the correct marginal coverage
(89.8\%). However, this marginal coverage is achieved by substantially
undercovering some classes and overcovering others. For example, $\mathsf{water\
jug}$ is severely undercovered: the prediction sets only include it in 50.8\% of
the cases where it is the true label. On the other hand, $\mathsf{flamingo}$ is
substantially overcovered: it achieves a class-conditional coverage of
99.2\%. This underscores the need for more refined methods in order to achieve
the class-conditional coverage defined in \eqref{eq:class-conditional-coverage}. 

\bigskip 

In principle, it is possible to achieve \eqref{eq:class-conditional-coverage} by  
splitting the calibration data by class and running conformal prediction once
for each class \citep{vovk2012conditional}. We refer to this as $\classwise$.
However, this procedure fails to be useful in many real applications since
data-splitting can result in very few calibration examples for each class-level
conformal prediction procedure. This typically happens in problem settings where
we have many classes but limited data; in such settings, the classwise procedure 
tends to be overly conservative and produces prediction sets that are too large
to be practically useful. We note that previous papers
\citep{shi2013applications, lofstrom2015bias} on class-conditional conformal
have not focused on the many classes regime and have instead studied binary or
at most 10-way classification tasks. 

\begin{figure}[h]
\centering
\includegraphics[width=0.95\textwidth]{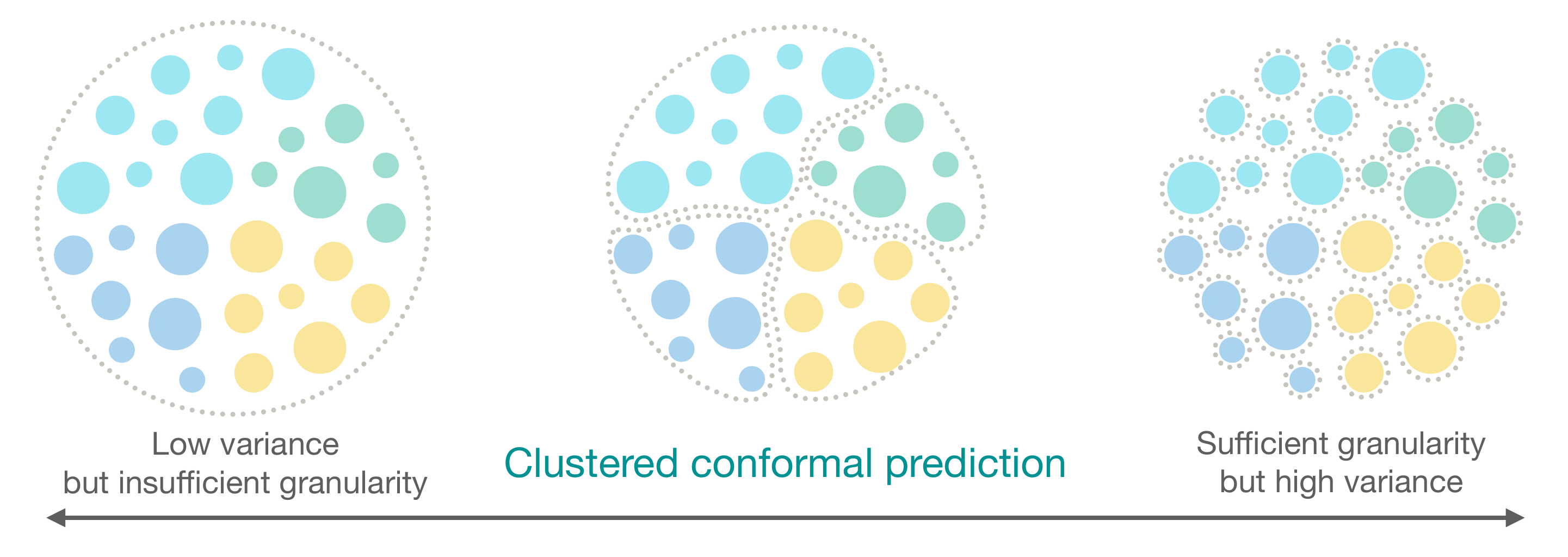} 
\caption{A schematic comparison of conformal prediction methods, including the
  $\clustered$ method we propose. Each colored circle represents the calibration
  data for a particular class. Existing methods fall on extremes of the spectrum:
  $\standard$ is very stable because it groups data for all classes together,
  but it is not able to treat classes differently when needed; $\classwise$ has
  high granularity, but it splits all of the data by class and consequently has
  high variance in limited-data settings; $\clustered$ strikes a balance by
  grouping together data for ``similar'' classes.}  
\label{fig:method_comparison_diagram}
\vspace{15pt} 
\end{figure}

In this work, we focus on the challenging limited-data, many-class
classification setting and we develop a method targeted at class-conditional  
coverage. Our method mitigates the problems that arise from data-splitting by
clustering together classes that have similar score distributions and combining
the calibration data for those classes. Figure
\ref{fig:method_comparison_diagram} illustrates how the method we propose
strikes a balance between $\standard$ and $\classwise$. As we will later show in
our experiments, this can improve class-conditional coverage in many situations.     

\paragraph{Our contributions.}
We make three advances in tackling the problem of class-conditional coverage.
\begin{itemize}
\item We propose an extension of conformal prediction called \textit{clustered
    conformal prediction} that often outperforms standard and classwise
  conformal, in terms of class-conditional coverage, when there is limited
  calibration data available per class.    

\item We present a comprehensive empirical evaluation of class-conditional 
  coverage for conformal methods on four large-scale classification data sets,
  each with many (100 to 1000) classes.   

\item We provide general guidelines to practitioners for how to choose an
  appropriate conformal prediction method in order to achieve class-conditional
  coverage in the problem setting at hand.    
\end{itemize} 

\subsection{Related work}

\paragraph{Mondrian conformal prediction.} 

We work in the split-conformal prediction framework
\citep{papadopoulos2002inductive, lei2018distribution} in which the training
data set (to train the base classifier) and calibration data set are
disjoint. Mondrian conformal prediction (MCP) is a general procedure that
encompasses many kinds of conditional conformal prediction
\citep{vovk2005algorithmic}. 
For any chosen grouping function $G: \cX \times \cY \to \mathcal{G}$ where $\mathcal{G}$ denotes the set of all groups, MCP provides coverage guarantees of the form $\P(\Ytest \in \mathcal{C}(\Xtest) \mid G(\Xtest, \Ytest) = g) \geq 1-\alpha$ for all groups $g \in \mathcal{G}$. 
The high-level idea behind MCP is to split the calibration data
by group and then run conformal prediction on
each group. The $\classwise$ conformal procedure is a special case of MCP where
each value of $\cY$ defines a group (this is also sometimes referred to as 
label-conditional conformal prediction).

To the best of our knowledge, previous work has not focused on class-conditional
coverage of conformal methods in the many-classes, limited-data regime that is
common to many real classification applications. \citet{lofstrom2015bias}
present an empirical study of $\standard$ and $\classwise$ on binary, 3-way, and
4-way classification data sets (with a focus on the binary setting). For
class-imbalanced problems, they find that $\standard$ tends to overcover the
majority class, and undercover the minority class. \citet{shi2013applications}
consider $\classwise$ on the MNIST and USPS data sets, which are 10-way
classification problems. \citet{sun2017applying} use a cross-fold variant of
$\classwise$ for binary classification on imbalanced bioactivity
data. \citet{hechtlinger2018cautious} run class-conditional experiments on the
3-class Iris data set, and provide preliminary ideas on how to incorporate
interactions between classes using density estimation.
\citet{guan2022prediction} use $\classwise$ in an outlier detection
context. 
\cite{sadinle2019least} consider $\classwise$ with a modification to avoid empty prediction sets and perform experiments on 3-way or 10-way classification tasks.
To reiterate, the aforementioned papers all focus on the regime that is
generally favorable to $\classwise$, where data is abundant relative to the
number of classes. Our work focuses on the limited-data regime that often arises
in practice.

\paragraph{Other types of conditional conformal prediction.} 

``Conditional coverage'' is a term that, within the conformal prediction
literature, often refers to coverage at a particular input value $X=x$. It is known to be
impossible to achieve coverage conditional on $X=x$, for all $x$,
without invoking further distributional assumptions \citep{lei2014distribution, 
  vovk2012conditional, barber2021limits}. However, approximate $X$-conditional 
coverage can be achieved in practice by designing better score functions
\citep{romano2019conformalized, romano2020classification,
  angelopoulos2022image} or by modifying the conformal procedure itself
\citep{romano2020malice, guan2023localized, gibbs2023conformal}. Our work draws inspiration from the
latter camp, but our aim is to achieve class-conditional ($Y$-conditional)
coverage rather than $X$-conditional coverage. Note that $X$-conditional  
coverage is hard to interpret in settings in which $Y$ is not random given $X=x$ 
(e.g., in image classification, if $x$ is an image of a dog, then the true label is always $y=\mathsf{dog}$, with no intrinsic randomness after conditioning on $X=x$). In such
settings, it is more natural to consider class-conditional coverage.        

\subsection{Preliminaries} 
\label{sec:definitions}

We work in the classification setting in which each input $X_i \in \cX$ has a
class label $Y_i \in \cY$, for some discrete set $\cY$. Let \smash{$\{(X_i,
  Y_i)\}_{i=1}^N$} denote a \emph{calibration data set}, where each
\smash{$(X_i, Y_i) \overset{\mathrm{iid}}{\sim} F$}. Given a new independent
test point $(\Xtest, \Ytest) \sim F$, our goal is to construct (without
knowledge of $\Ytest$) a prediction set $\cC(\Xtest)$ that satisfies 
\eqref{eq:class-conditional-coverage} for some user-specified $\alpha \in
[0,1]$.     

Let $s: \cX \times \cY \to \R$ denote a \emph{(conformal) score function},
where we take $s(x,y)$ to be negatively oriented, which means that lower scores 
indicate a better agreement between the input $x$ and the proposed class label 
$y$. The score function is typically derived from a pre-trained classifier 
\smash{$\fhat$} (a simple example to keep in mind is \smash{$s(x,y) = 1 -
  \fhat_y(x)$}, where \smash{$\fhat_y(x)$} represents the \smash{$y\th$} entry
of the softmax vector output by \smash{$\fhat$} for the input $x$). For brevity,
we denote the score of the \smash{$i\th$} calibration data point as $s_i =
s(X_i, Y_i)$. For $\tau \in [0,1]$ and a finite set $A \subseteq \R$, let 
$\quantile(\tau, A)$ denote the smallest $a \in A$ such that $\tau$ fraction of
elements in $A$ are less than or equal to $a$. For $\tau>1$, we take
$\quantile(\tau, A) = \infty$.   

With this notation, the $\standard$ conformal prediction sets are given by  
\[
\cC_{\standard}(\Xtest) = \{y: s(\Xtest, y) \leq \qhat\}, 
\]
where
\[
\qhat = \quantile\bigg( \frac{\lceil(N+1)(1-\alpha)\rceil}{N}, \{s_i\}_{i=1}^N
\bigg). 
\]
These prediction sets are guaranteed to satisfy \eqref{eq:marginal-coverage}
\citep{vovk2005algorithmic}. We can interpret \smash{$\qhat$} as a finite-sample
adjusted $(1-\alpha)$-quantile of the scores in the calibration data set. Note 
that all $N$ data points are used for computing a single number,
\smash{$\qhat$}. 

In contrast, the $\classwise$ procedure computes a separate quantile for every
class. Let $\cI^y = \{i \in [N]: Y_i = y\}$ be the indices of examples in the
calibration data set that have label $y$. The $\classwise$ conformal prediction
sets are given by 
\[
\cC_{\classwise}(\Xtest) = \{y: s(\Xtest, y) \leq \qhat^y\},
\]
where
\[
\qhat^y = \quantile\bigg( \frac{\lceil(|\cI^y|+1)(1-\alpha)\rceil}{|\cI^y|},
  \{s_i\}_{i \in \cI^y} \bigg). 
\]
These prediction sets are guaranteed to satisfy
\eqref{eq:class-conditional-coverage} \citep{vovk2012conditional}. 
However, the quantile \smash{$\qhat^y$} is only computed using a subset of the
data of size $|\mathcal{I}_y|$, which might be quite
small. Importantly, for any class $y$ for which $|\cI^y| < (1/\alpha) - 1$, we
will have \smash{$\qhat^y = \infty$}, hence any prediction set generated by
$\classwise$ will include $y$, no matter the values of the conformal scores.    

The advantage of $\standard$ is that we do not need to split up the calibration
data to estimate \smash{$\qhat$}, so the estimated quantile has little noise
even in limited-data settings; however, it does not in general yield
class-conditional coverage. Conversely, $\classwise$ is guaranteed to achieve
class-conditional coverage, since we estimate a different threshold
\smash{$\qhat^y$} for every class; however, these estimated quantiles can be
very noisy due to the limited data, leading to erratic behavior, such as large
sets.  

When we assume the scores are almost surely distinct, the exact distribution of
the $\classwise$ class-conditional coverage of class $y$ given a fixed
calibration set is \citep{vovk2012conditional, angelopoulos2023gentle}:      
\[
\P\big( \Ytest \in \cC_{\classwise}(\Xtest) \mid \Ytest = y, \, \{(X_i,
Y_i)\}_{i=1}^N \big) \sim \mathrm{Beta}( k_\alpha^y, |\cI^y|+1 - k_\alpha^y), 
\] 
where \smash{$k_\alpha^y = \lceil(|\cI^y|+1)(1-\alpha)\rceil$}. Note that the
probability on the left-hand side above is taken with respect to the test point $\Xtest$ as the only source of randomness, as we have conditioned
on the calibration set. The beta distribution \smash{$\mathrm{Beta}( k_\alpha^y, 
  |\cI^y|+1 - k_\alpha^y)$} has mean \smash{$k_\alpha^y / (|\cI^y|+1)$} (which,
as expected, is always at least $1-\alpha$) and variance    
\[
\frac{k_\alpha^y (|\cI^y|+1 - k_\alpha^y)}{(|\cI^y|+1)^2 (|\cI^y|+2)} \approx
\frac{\alpha (1-\alpha)}{|\cI^y|+2},
\]
which can be large when $|\cI^y|$ is small. For example, if class $y$ only has
10 calibration examples and we seek 90\% coverage, then the class-conditional
coverage given a fixed calibration set is distributed as $\mathrm{Beta}(10,1)$,
so there is probability $\approx 0.107$ that the coverage of class $y$ will be
less than 80\%. Somewhat paradoxically, the variance of the class-conditional
coverage means that on a given realization of the calibration set, the
$\classwise$ method can exhibit poor coverage on a substantial fraction of
classes if the number of calibration data points per class is limited.

\section{Clustered conformal prediction} 
\label{sec:clustered_conformal}
With the goal of achieving the class-conditional coverage guarantee in
\eqref{eq:class-conditional-coverage}, we introduce \emph{clustered conformal
  prediction}. Our method strikes a balance between the granularity of
$\classwise$ and the data-pooling of $\standard$ by grouping together
classes according to a clustering function. For each cluster, we calculate a
single quantile based on all data in that cluster. We design the
clustering algorithm so that clustered classes have similar score distributions,
and therefore, the resulting cluster-level quantile is applicable to all classes
in the cluster. Next, in Section \ref{sec:meta_algorithm}, we formally describe
the clustered conformal prediction method; then, in Section
\ref{sec:clustering}, we describe the clustering step in detail. 

\subsection{Meta-algorithm} 
\label{sec:meta_algorithm}

To begin, we randomly split the calibration data set into two parts: the
\emph{clustering data set} $D_1 = \{(X_i,Y_i): i \in \cI_1\}$, for performing
clustering, and a \emph{proper calibration data set} $D_2 = \{(X_i,Y_i): i \in
\cI_2\}$, for computing the conformal quantiles, where \smash{$|\cI_1| =
  \lfloor\gamma N\rfloor$} and $|\cI_2| = N-|\cI_1|$ for some tuning parameter
$\gamma \in [0,1]$. Then, we apply a clustering algorithm to $D_1$ to obtain a 
\emph{clustering function} \smash{$\hhat : \cY \to \{1,\dots,M\} \cup
  \{\nullclust\}$} that maps each class $y \in \cY$ to one of $M$ clusters or
the ``null'' cluster (denoted $\nullclust$). The reason that we include the null 
cluster is to handle rare classes that do not have sufficient data to be
confidently clustered into any of the $M$ clusters. Details on how to create 
\smash{$\hhat$} are given in the next subsection. 

After assigning classes to clusters, we perform the usual conformal calibration procedure within each cluster. Denote by \smash{$\cI_2^y = \{i \in \cI_2 : Y_i=y\}$} the 
indices of examples in $D_2$ with label $y$, and by \smash{$\cI_2(m) = \cup_{y
  : \hhat(y)=m} \, \cI_2^y$} the indices of examples in $D_2$ with labels in
cluster $m$. The $\clustered$ conformal prediction sets are given by
\[
\cC_{\clustered}(\Xtest) = \{y: s(\Xtest, y) \leq \qhat(\hhat(y))\},
\]
where, for $m=1,\dots,M$,
\[
\qhat(m) = \quantile\bigg(
\frac{\lceil(|\cI_2(m)|+1)(1-\alpha)\rceil}{|\cI_2(m)|}, \{s_i\}_{i \in
  \cI_2(m)} \bigg),
\]
and
\[
\qhat(\nullclust) = \quantile\bigg(
\frac{\lceil(|\cI_2|+1)(1-\alpha)\rceil}{|\cI_2|}, \{s_i\}_{i \in \cI_2}
\bigg). 
\]
In words, for each cluster $m$, we group together examples for all classes in
that cluster and then estimate a cluster-level quantile \smash{$\qhat(m)$}. 
When constructing prediction sets, we include class $y$ in the set if the
conformal score for class $y$ is less than or equal to the quantile for the
cluster that contains $y$. For classes assigned to the null cluster, we compare
the conformal score against the quantile that would be obtained from running 
$\standard$ on the proper calibration set.

We now consider the properties of the $\clustered$ prediction sets. For all
classes that do not belong to the null cluster, we have the following guarantee
(the proof of this result, and all other proofs, are deferred to Appendix
\ref{sec:proofs}).  

\begin{proposition} 
\label{prop:cluster-conditional-coverage}
The prediction sets $\cC = \cC_{\clustered}$ from $\clustered$ achieve
cluster-conditional coverage:  
\begin{equation}
\label{eq:cluster-conditional-coverage}
\P(\Ytest \in \cC(\Xtest) \mid \hhat(\Ytest)=m) \geq 1- \alpha, \quad \text{for
  all clusters } m=1,\dots,M.  
\end{equation}
\end{proposition}

This coverage result comes from the exchangeability between a test point \emph{drawn from cluster $m$} and all of the calibration points that belong to cluster $m$. We get this exchangeability for free from the assumption that the calibration point and test points are sampled i.i.d.\ from the same distribution.
Cluster-conditional coverage is a stronger guarantee than marginal coverage, but
it is still not as strong as the class-conditional coverage property that we
hope to achieve. However, cluster-conditional coverage implies class-conditional
in an idealized setting: suppose we have access to an ``oracle'' clustering
function $h^*$ that produces $M$ clusters such that, for each cluster
$m=1,\dots,M$, the scores for all classes in this cluster are exchangeable (this would hold, for example, if all classes in the same cluster have the same score distribution). 
In this case, we have a guarantee on class-conditional coverage.         

\begin{proposition} 
\label{prop:class-conditional-coverage}
If \smash{$\hhat = h^*$}, the ``oracle'' clustering function as described above, 
then the prediction sets from $\clustered$ satisfy class-conditional coverage
\eqref{eq:class-conditional-coverage} for all classes $y$ such that $h^*(y)
\neq \nullclust$.  
\end{proposition}

This coverage result arises because the oracle clustering function ensures exchangeability between a test point \emph{drawn from class $y$} and all of the calibration points that belong to the cluster to which $y$ belongs. To try to achieve this exchangeability, we need to construct the clusters carefully. Specifically, we want to design \smash{$\hhat$} so that the scores within
each cluster $m=1,\dots,M$ are as close to identically distributed as possible, an idea 
we pursue next. 

\subsection{Quantile-based clustering} 
\label{sec:clustering}

We seek to cluster together classes that have similar score distributions. To do
so, we first summarize the empirical score distribution for each class via a
vector of score quantiles evaluated at a discrete set of levels $\tau \in \cT
\subseteq [0,1]$. In this embedding space, a larger distance between classes 
means their score distributions are more different. 
While there are various options for defining such an embedding, recall that to
achieve class-conditional coverage, we want to accurately estimate the
(finite-sample adjusted) $(1-\alpha)$-quantile for the score distribution of
each class. Thus, we want to group together classes with similar quantiles,
which is what our embedding is designed to facilitate. After obtaining these
embeddings, we can then simply apply any clustering algorithm, such as
$k$-means.   

In more detail, denote by \smash{$\cI_1^y = \{i \in \cI_1 : Y_i = y\}$} the
indices of examples in $D_1$ with label $y$. We compute quantiles of the scores
\smash{$\{s_i\}_{i \in \cI_1^y}$} from class $y$ at the levels 
\[
\cT = \bigg\{ \frac{\lceil(|\cI^y|+1) \tau \rceil}{|\cI^y|} : \tau \in \{0.5,
0.6, 0.7, 0.8, 0.9\} \cup \{1-\alpha\} \bigg\}.
\]
and collect them into an embedding vector $z^y \in \R^{|\cT|}$. If
\smash{$|\cI_1^y| < n_\alpha$}, where $n_\alpha = (1/\min\{\alpha, 0.1\})-1$,
then the uppermost quantile in $z^y$ will not be finite, so we simply assign
$y$ to the null cluster. For a pre-chosen number of clusters $M$, we run
$k$-means clustering with $k=M$ on the data \smash{$\{z^y\}_{y \in \cY \setminus
    \cY_{\nullclust}}$}, where \smash{$\cY_{\nullclust}$} denotes the set of
labels assigned to the null cluster. More specifically, we use a weighted version of
$k$-means where the weight for class $y$ is set to \smash{$|\cI_1^y|^{1/2}$};
this allows the class embeddings computed from more data to have more influence
on the cluster centroids. We denote the cluster mapping that results from this
procedure by \smash{$\hhat : \cY \to \{1,\dots,M\} \cup \{\nullclust\}$}.

Of course, we cannot generally recover an oracle clustering function $h^*$ with
the above (or any practical) procedure, so Proposition
\ref{eq:class-conditional-coverage} does not apply. However, if the score
distributions for the classes that \smash{$\hhat$} assigns to the same cluster
are similar enough, then we can
provide an approximate class-conditional coverage guarantee. We measure similarity in terms of the Kolmogorov-Smirnov (KS) distance, which is defined between random variables $X$ and $Y$ as \smash{$D_{\mathrm{KS}}(X, Y) = \sup_{\lambda \in \R} \, |\P(X \leq \lambda) - \P(Y\leq \lambda)|$}.

\begin{proposition} 
\label{prop:approx-class-conditional-coverage}
Let $S^y$ denote a random variable sampled from the score distribution for class
$y$, and assume that the clustering map \smash{$\hhat$} satisfies  
\[
D_{\mathrm{KS}}(S^y, S^{y'}) \leq \epsilon, \quad \text{for all $y, y'$ such that $\hhat(y)
  = \hhat(y') \neq \nullclust$}.
\]
Then, for $\cC = \cC_{\clustered}$ and for all classes $y$ such that such that
\smash{$\hhat(y) \neq \nullclust$}, 
\[
\P(\Ytest \in \cC(\Xtest) \mid \Ytest = y) \geq 1- \alpha - \epsilon.
\]
\end{proposition}

To summarize, the two main tuning parameters used in the proposed method are
$\gamma \in [0,1]$, the fraction of points to use for the clustering step, and
$M \geq 1$, the number of clusters to use in $k$-means. While there are no
universal fixed values of these parameters that serve all applications equally
well, we find that simple heuristics for setting these parameters often work 
well in practice, which we describe in the next section.  


\section{Experiments}
\label{sec:experiments}
We evaluate the class-conditional coverage of $\standard$, $\classwise$, and
$\clustered$ conformal prediction on four large-scale image data sets using
three conformal score functions on each. Code for reproducing our experiments is 
available at \url{https://github.com/tiffanyding/class-conditional-conformal}.

\subsection{Experimental setup} 
\label{sec:experimental_setup}

We run experiments on the ImageNet \citep{russakovsky2015imagenet}, CIFAR-100 \citep{krizhevsky09learningmultiple}, Places365 \citep{zhou2018places}, and iNaturalist \citep{van2018inaturalist} image
classification data sets, whose characteristics are summarized in Table
\ref{tab:dataset_summary}. The first three have roughly balanced classes, the
fourth, iNaturalist, has high class imbalance. We consider three conformal score
functions: $\softmax$, one minus the softmax output of the base classifier;
$\APS$, a score designed to improve $X$-conditional coverage;   
and $\RAPS$, a regularized version of $\APS$ that often produces smaller sets.
Precise definitions of the score functions are provided in Appendix
\ref{sec:score_functions}; we refer also to \citet{romano2020classification,
  angelopoulos2021uncertainty} for the details and motivation behind $\APS$ and
$\RAPS$. Throughout, we set $\alpha=0.1$ for a desired coverage level of 90\%.   

\begin{table}[h!]
\caption{Description of data sets. The class balance metric is described
  precisely in Appendix \ref{sec:class_balance_metric}.}  
\label{tab:dataset_summary}
\centering 
\begin{tabular}{l c c c c} 
\emph{Data set} \hspace{-10pt} & ImageNet & CIFAR-100 & Places365 & iNaturalist \\
\midrule
\emph{Number of classes} & 1000 & 100 & 365 & 663$^*$ \\ 
\emph{Class balance} & 0.79 & 0.90 & 0.77 & 0.12 \\
\emph{Example classes} & mitten  & orchid & beach & salamander \\ 
& triceratops & forest & sushi bar & legume \\
& guacamole & bicycle & catacomb & common fern \\
\end{tabular}
\begin{tablenotes}
\footnotesize 
\item[] *The number of classes in the iNaturalist data set can be adjusted by   
  selecting which taxonomy level (e.g., species, genus, family) to use as the
  class labels. We use the species family as our label and then filter out any
  classes with $<250$ examples in order to have sufficient examples to properly    
  perform evaluation.
\end{tablenotes}
\end{table}

Our experiments all follow a common template. First, we fine-tune a
pre-trained neural network as our base classifier (for details on model
architectures, see Appendix \ref{sec:model_training}) on a small subset
\smash{$D_{\mathrm{fine}}$} of the original data, leaving the rest for
calibration and validation purposes. 
We construct calibration sets of varying size by changing the average number of points in each class, denoted $\navg$.
For each $\navg \in \{10, 20, 30, 40, 50,
75, 100, 150\}$, we construct a calibration set \smash{$D_{\mathrm{cal}}$} by
sampling $\navg \times |\cY|$ examples without replacement from the remaining
data \smash{$D_{\mathrm{fine}}^c$} (where $^c$ denotes the set complement). We estimate the conformal quantiles for 
$\standard$, $\classwise$, and $\clustered$ on $D_{\mathrm{cal}}$. The remaining
data \smash{$(D_{\mathrm{fine}} \cup D_{\mathrm{cal}})^c$} is used as the
validation set for computing coverage and set size metrics. Finally, this
process---splitting \smash{$D_{\mathrm{fine}}^c$} into random calibration and
validation sets---is repeated ten times, and the reported metrics are averaged
over these repetitions. 

\paragraph{Details about clustering.}

For $\clustered$, we choose $\gamma \in [0,1]$ (the fraction of calibration
data points used for clustering) and $M \geq 1$ (the number of clusters) in the
following way. First, we define \smash{$n_{\min} = \min_{y \in \cY} |\cI^y|$},
the number of examples in the rarest class in the calibration set, and 
$n_{\alpha} = (1/\alpha)-1$, the minimum sample size needed so that the
finite-sample adjusted $(1-\alpha)$-quantile used in conformal prediction is
finite (e.g., $n_{\alpha}=9$ when $\alpha=0.1$). Now define \smash{$\tilde{n} = 
  \max(n_{\min}, n_{\alpha})$} and let $K$ be the number of classes with at
least \smash{$\tilde{n}$} examples.  We then set $\gamma = K/(75+K)$ and
\smash{$M = \lfloor \gamma \tilde{n} / 2 \rfloor$}. These choices are motivated
by two goals: we want $M$ and $\gamma$ to grow together (to find more clusters,
we need more samples for clustering), and we want the proper calibration set to
have at least 150 points per cluster on average; see Appendix
\ref{sec:clustering_parameters} for details on how the latter is
achieved. Clustering is carried out by running $k$-means on the quantile-based
embeddings, as described in Section \ref{sec:clustering}; we use the
implementation in \verb|sklearn.cluster.KMeans|, with the default settings
\citep{scikit-learn}.

\subsection{Evaluation metrics}

Denote the validation data set by \smash{$\{(X'_i,Y'_i)\}_{i=1}^{N'}$} (recall
this is separate from the fine-tuning and calibration data sets) and let
\smash{$\cJ^y =  \{i \in [N'] : Y'_i= y\}$} be the indices of validation
examples with label $y$. For any given conformal method, we define
\smash{$\hat{c}_y = \frac{1}{|\cJ^y|} \sum_{i \in \cJ^y} \ind{Y'_i \in
    \cC(X'_i)}$} as the empirical class-conditional coverage of class $y$. Our
main evaluation metric is the \emph{average class coverage gap} (CovGap):         
\[
\text{CovGap} = 100 \times \frac{1}{|\cY|}\sum_{y \in \cY} |\hat{c}_y -
(1-\alpha)|.  
\]
This measures how far the class-conditional coverage is from the desired
coverage level of $1-\alpha$, in terms of the $\ell_1$ distance across all
classes (multiplied by 100 to put it on the percentage scale). We also measure
the sharpness of the prediction sets by \emph{average set size} (AvgSize):   
\[
\text{AvgSize} = \frac{1}{N'}\sum_{i=1}^{N'} |\cC(X'_i)|.
\]
For a given class-conditional coverage level (a given CovGap), we want a smaller
average set size.  

\subsection{Results}


\begin{figure}[b!] 
\centering
\includegraphics[width=0.965\textwidth]{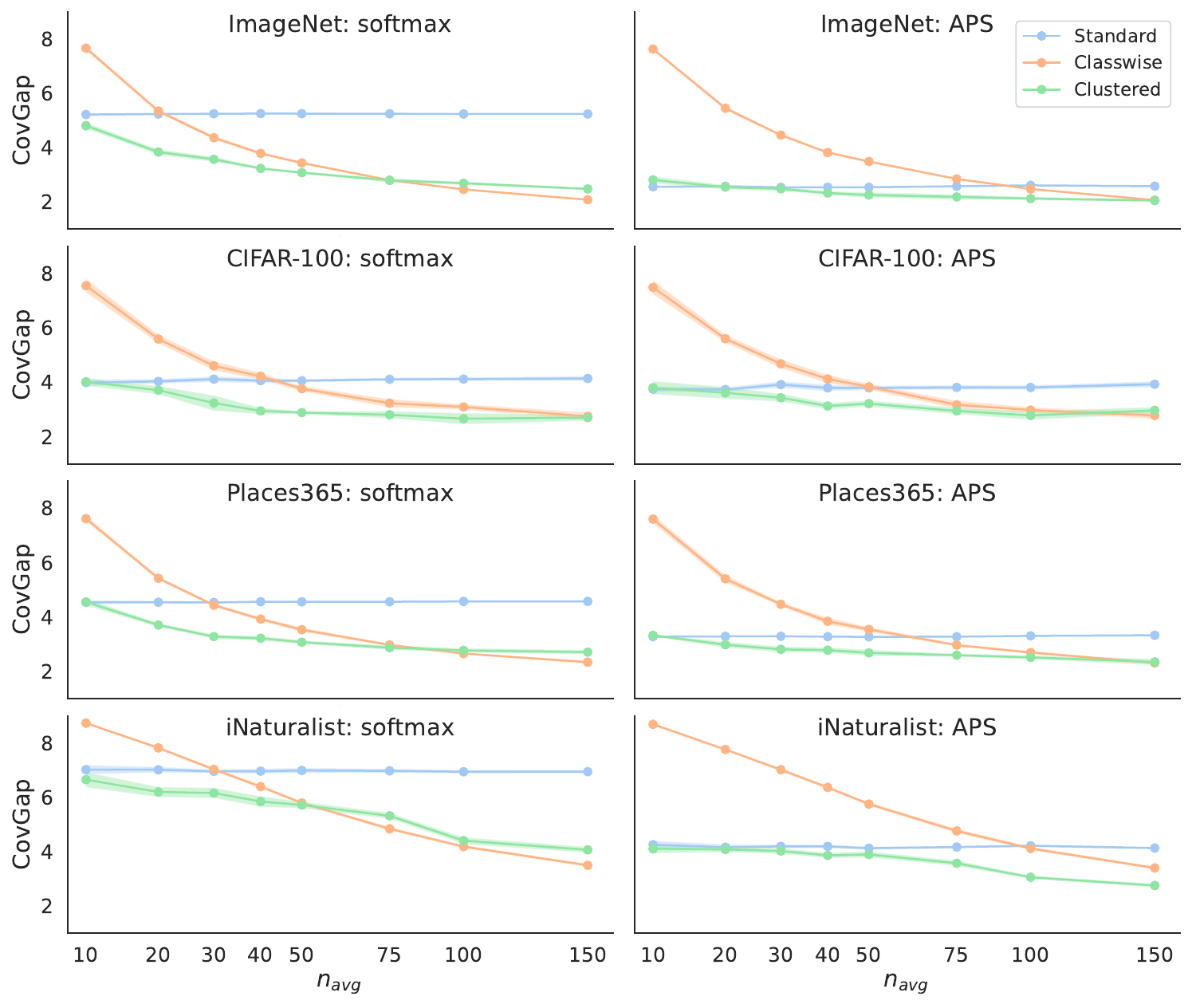} 
\caption{Average class coverage gap for ImageNet, CIFAR-100, Places365, and
  iNaturalist, for the $\softmax$ (left) and $\APS$ (right) scores, as we vary
  the average number of calibration examples per class. The shaded regions
  denote $\pm 1.96$ times the standard errors (often, the standard errors are
  too small to be visible).}   
\label{fig:alldatasets_covgap}
\vspace{-10pt}
\end{figure}

To begin, we investigate the CovGap of the methods on each data set, and display 
the results in Figure \ref{fig:alldatasets_covgap}. In brief, $\clustered$
achieves the best or comparable performance across all settings. Restricting our
attention to the baseline methods, note that the CovGap of $\standard$ does not
change much as we vary $\navg$; this is as expected, because $\navg \times
|\cY|$ samples are being used to estimate the conformal quantile
\smash{$\qhat$}, which will be stable regardless of $\navg$, provided $|\cY|$ is
large. Conversely, the CovGap of $\classwise$ decreases significantly as $\navg$
increases, because the classwise conformal quantiles 
\smash{$\qhat^y$} are volatile for small $\navg$.   

For the $\softmax$ score (left column of the figure), we see that
$\clustered$ clearly outperforms $\standard$, and the gap widens with $\navg$;
further, $\clustered$ outperforms $\classwise$ for small enough values of
$\navg$ (that is, $\navg < 75$ for ImageNet and Places365, $\navg < 150$ for  
CIFAR-100, and $\navg < 50$ for iNaturalist). The comparison between
$\clustered$ and $\classwise$ is qualitatively similar for $\APS$ score (right
column of the figure), with the former outperforming the latter for small enough
values of $\navg$. However, the behavior of $\standard$ changes notably as we
move from $\softmax$ to $\APS$: it becomes comparable to $\clustered$ and only
slightly worse for large $\navg$. Lastly, the results for $\RAPS$ (not shown, and
deferred to Appendix \ref{sec:results_other_score_functions}) are similar to
$\APS$ but the CovGap is shifted slightly higher. 

To examine the potential tradeoffs between class-conditional coverage and
average set size, we focus on the $\navg \in \{10, 20, 50, 75\}$ settings and
report CovGap and AvgSize in Table \ref{tab:results} for all data sets and score
functions. We see that $\clustered$ achieves the best or near-best CovGap in any
experimental combination, and its improvement over the baselines is particularly
notable in the regime where data is limited, but not extremely limited ($\navg =
20$ or 50). Meanwhile, AvgSize for $\standard$ and $\clustered$ is relatively
stable across values of $\navg$, with the latter being generally slightly
larger; AvgSize for $\classwise$ decreases as we increase $\navg$, but is still
quite large relative to $\standard$ and $\clustered$, especially for iNaturalist. 

We finish with two more remarks. First, we find CovGap tends to behave quite 
similarly to the fraction of classes that are drastically undercovered, which we
define as having a class-conditional coverage at least 10\% below the desired 
level. Results for this metric are given in Appendix
\ref{sec:additional_metrics}. Second, as is the case in any conformal
method, auxiliary randomization can be applied to $\standard$, $\classwise$, or
$\clustered$ in order to achieve a coverage guarantee (marginal,
class-conditional, or cluster-conditional, respectively) of \emph{exactly}
$1-\alpha$, rather than \emph{at least} $1-\alpha$. These results are
included in Appendix \ref{sec:randomized_methods}. In terms of CovGap, we find
that randomization generally improves $\classwise$, and does not change
$\standard$ or $\clustered$ much at all; however, the set sizes from randomized
$\classwise$ are still overall too large to be practically useful (moreover,
injecting auxiliary randomness is prediction sets is not always practically
palatable).    

\begin{table}[h!]
\scriptsize 
\setlength{\tabcolsep}{3pt} 
\caption{Average class coverage gap and average set size for select values of
  $\navg$ on ImageNet, CIFAR-100, Places365, and iNaturalist. Bold emphasizes
  the best (within $\pm 0.2$) class coverage gap in each experimental
  combination. Standard errors are reported in parentheses.}     
\label{tab:results}
\centering 
\begin{tabular}{lllcrcrcrcr}
\toprule
            &         &              & \multicolumn{2}{c}{$n_{\mathrm{avg}}=10$} & \multicolumn{2}{c}{$n_{\mathrm{avg}}=20$} & \multicolumn{2}{c}{$n_{\mathrm{avg}}=50$} & \multicolumn{2}{c}{$n_{\mathrm{avg}}=75$} \\ \cmidrule(lr){4-5}\cmidrule(lr){6-7} \cmidrule(lr){8-9} \cmidrule(lr){10-11}\
            &         &              &              CovGap &      AvgSize &              CovGap &      AvgSize &              CovGap &      AvgSize &              CovGap &     AvgSize \\
Data set & Score & Method &                     &              &                     &              &                     &              &                     &             \\
\midrule
ImageNet & $\softmax$ & $\standard$ &           5.2 (0.0) &    1.9 (0.0) &           5.2 (0.0) &    1.9 (0.0) &           5.3 (0.0) &    1.9 (0.0) &           5.2 (0.0) &   1.9 (0.0) \\
            &         & $\classwise$ &           7.7 (0.0) &  354.4 (2.0) &           5.4 (0.0) &   23.0 (0.9) &           3.4 (0.0) &    5.1 (0.1) &           \attn{2.8} (0.0) &   4.2 (0.1) \\
            &         & $\clustered$ &           \attn{4.9} (0.1) &    2.5 (0.1) &           \attn{3.9} (0.0) &    2.7 (0.1) &           \attn{3.1} (0.0) &    2.6 (0.1) &           \attn{2.8} (0.0) &   2.7 (0.0) \\
            \cline{3-11} 
            & $\APS$ & $\standard$ &           \attn{2.6} (0.0) &   25.9 (0.3) &           \attn{2.6} (0.0) &   25.8 (0.1) &           \attn{2.5} (0.0) &   25.6 (0.1) &           2.6 (0.0) &  25.7 (0.1) \\
            &         & $\classwise$ &           7.6 (0.0) &  394.2 (1.9) &           5.5 (0.0) &   76.9 (0.8) &           3.5 (0.0) &   39.7 (0.2) &           2.9 (0.0) &  35.0 (0.2) \\
            &         & $\clustered$ &           3.0 (0.1) &   29.6 (1.5) &           \attn{2.6} (0.0) &   27.0 (0.7) &           \attn{2.3} (0.0) &   27.2 (0.5) &           \attn{2.2} (0.0) &  27.3 (0.4) \\
            \cline{3-11} & $\RAPS$ & $\standard$ &           \attn{3.0} (0.0) &    5.2 (0.0) &           \attn{3.0} (0.0) &    5.1 (0.0) &           3.0 (0.0) &    5.1 (0.0) &           3.0 (0.0) &   5.1 (0.0) \\
            &         & $\classwise$ &           7.7 (0.1) &  361.9 (1.9) &           5.6 (0.0) &   34.2 (1.0) &           3.4 (0.0) &    8.8 (0.3) &           2.8 (0.0) &   7.3 (0.1) \\
            &         & $\clustered$ &           \attn{3.1} (0.0) &    7.7 (1.0) &           \attn{2.9} (0.0) &    6.6 (0.6) &           \attn{2.6} (0.0) &    6.5 (0.3) &           \attn{2.4} (0.0) &   6.8 (0.3) \\
\midrule
CIFAR-100 & $\softmax$ & $\standard$ &           \attn{4.0} (0.1) &    8.1 (0.2) &           4.0 (0.0) &    7.9 (0.2) &           4.1 (0.0) &    7.9 (0.1) &           4.1 (0.0) &   8.0 (0.1) \\
            &         & $\classwise$ &           7.6 (0.1) &   47.2 (0.6) &           5.6 (0.1) &   19.3 (0.5) &           3.8 (0.0) &   11.9 (0.2) &           3.2 (0.1) &  10.8 (0.1) \\
            &         & $\clustered$ &           \attn{4.2} (0.1) &    8.9 (0.5) &           \attn{3.6} (0.1) &    8.9 (0.3) &           \attn{2.9} (0.1) &    9.3 (0.3) &           \attn{2.8} (0.0) &   9.1 (0.2) \\
            \cline{3-11} & $\APS$ & $\standard$ &           \attn{3.7} (0.1) &   11.2 (0.3) &           \attn{3.7} (0.0) &   10.8 (0.2) &           3.8 (0.0) &   11.0 (0.1) &           3.8 (0.0) &  11.0 (0.1) \\
            &         & $\classwise$ &           7.5 (0.1) &   49.4 (0.5) &           5.6 (0.1) &   22.6 (0.5) &           3.8 (0.1) &   15.0 (0.2) &           3.2 (0.1) &  13.7 (0.1) \\
            &         & $\clustered$ &           4.0 (0.1) &   11.9 (0.5) &           \attn{3.6} (0.1) &   12.1 (0.5) &           \attn{3.1} (0.1) &   12.2 (0.4) &           \attn{2.8} (0.1) &  12.2 (0.2) \\
            \cline{3-11} & $\RAPS$ & $\standard$ &           \attn{4.7} (0.1) &    8.6 (0.3) &           4.8 (0.1) &    8.3 (0.3) &           4.8 (0.0) &    8.3 (0.1) &           4.9 (0.1) &   8.1 (0.1) \\
            &         & $\classwise$ &           7.5 (0.1) &   47.5 (0.7) &           5.6 (0.1) &   20.7 (0.6) &           \attn{3.8} (0.1) &   13.3 (0.2) &           \attn{3.3} (0.1) &  12.1 (0.1) \\
            &         & $\clustered$ &           \attn{4.9} (0.1) &    9.5 (0.5) &           \attn{4.5} (0.1) &    8.7 (0.5) &           \attn{3.7} (0.1) &    9.5 (0.3) &           \attn{3.4} (0.1) &   9.9 (0.1) \\
\midrule
Places365 & $\softmax$ & $\standard$ &           \attn{4.5} (0.0) &    6.9 (0.1) &           4.5 (0.0) &    6.9 (0.0) &           4.6 (0.0) &    6.9 (0.0) &           4.6 (0.0) &   6.9 (0.0) \\
            &         & $\classwise$ &           7.6 (0.1) &  136.5 (1.7) &           5.4 (0.0) &   18.0 (0.2) &           3.5 (0.0) &   10.1 (0.1) &           \attn{3.0} (0.0) &   9.3 (0.1) \\
            &         & $\clustered$ &           \attn{4.5} (0.1) &    7.1 (0.1) &           \attn{3.9} (0.1) &    7.1 (0.1) &           \attn{3.0} (0.0) &    7.8 (0.1) &           \attn{2.9} (0.1) &   7.9 (0.1) \\
            \cline{3-11} & $\APS$ & $\standard$ &           \attn{3.3} (0.0) &   10.8 (0.1) &           \attn{3.3} (0.0) &   10.9 (0.1) &           3.3 (0.0) &   10.8 (0.1) &           3.3 (0.0) &  10.8 (0.1) \\
            &         & $\classwise$ &           7.6 (0.1) &  140.5 (1.7) &           5.4 (0.1) &   23.5 (0.3) &           3.5 (0.0) &   14.4 (0.1) &           3.0 (0.0) &  13.4 (0.1) \\
            &         & $\clustered$ &           \attn{3.5} (0.1) &   10.9 (0.3) &           \attn{3.1} (0.0) &   10.9 (0.2) &           \attn{2.6} (0.0) &   11.8 (0.1) &           \attn{2.7} (0.1) &  11.6 (0.2) \\
            \cline{3-11} & $\RAPS$ & $\standard$ &           \attn{3.8} (0.0) &    7.9 (0.0) &           3.9 (0.0) &    7.9 (0.0) &           3.8 (0.0) &    7.9 (0.0) &           3.8 (0.0) &   7.9 (0.0) \\
            &         & $\classwise$ &           7.6 (0.1) &  138.9 (1.7) &           5.5 (0.1) &   21.3 (0.3) &           3.6 (0.0) &   11.4 (0.1) &           \attn{3.0} (0.0) &  10.5 (0.1) \\
            &         & $\clustered$ &           \attn{3.9} (0.1) &    8.0 (0.1) &           \attn{3.4} (0.0) &    8.2 (0.2) &           \attn{3.0} (0.0) &    8.8 (0.1) &           \attn{2.9} (0.0) &   9.0 (0.2) \\
\midrule
iNaturalist & $\softmax$ & $\standard$ &           \attn{7.0} (0.1) &    3.1 (0.0) &           7.0 (0.1) &    3.1 (0.0) &           7.0 (0.1) &    3.2 (0.0) &           7.0 (0.0) &   3.2 (0.0) \\
            &         & $\classwise$ &           8.7 (0.0) &  469.4 (1.8) &           7.8 (0.0) &  364.0 (1.4) &           \attn{5.8} (0.0) &  148.7 (2.2) &           \attn{4.8} (0.0) &  55.3 (1.6) \\
            &         & $\clustered$ &           \attn{6.9} (0.2) &    3.1 (0.1) &           \attn{6.4} (0.1) &    3.4 (0.1) &           \attn{5.7} (0.1) &    3.7 (0.0) &           5.3 (0.1) &   3.8 (0.0) \\
            \cline{3-11} & $\APS$ & $\standard$ &           \attn{4.3} (0.1) &    8.3 (0.1) &           \attn{4.2} (0.1) &    8.4 (0.1) &           \attn{4.1} (0.0) &    8.3 (0.0) &           4.2 (0.0) &   8.3 (0.0) \\
            &         & $\classwise$ &           8.7 (0.0) &  472.8 (1.8) &           7.8 (0.0) &  368.4 (1.4) &           5.7 (0.0) &  153.9 (2.2) &           4.8 (0.0) &  60.7 (1.5) \\
            &         & $\clustered$ &           \attn{4.3} (0.1) &    8.1 (0.1) &           \attn{4.1} (0.1) &    8.3 (0.1) &           \attn{3.9} (0.0) &    8.5 (0.1) &           \attn{3.5} (0.0) &   8.7 (0.1) \\
            \cline{3-11} & $\RAPS$ & $\standard$ &           \attn{5.1} (0.1) &    5.1 (0.0) &           \attn{5.0} (0.0) &    5.2 (0.0) &           5.0 (0.0) &    5.1 (0.0) &           5.0 (0.0) &   5.2 (0.0) \\
            &         & $\classwise$ &           8.7 (0.0) &  473.8 (1.9) &           7.8 (0.0) &  369.2 (1.5) &           5.8 (0.0) &  155.3 (2.3) &           4.9 (0.0) &  63.1 (1.7) \\
            &         & $\clustered$ &           \attn{5.0} (0.1) &    5.0 (0.0) &           \attn{4.9} (0.1) &    5.2 (0.0) &           \attn{4.5} (0.1) &    5.4 (0.1) &           \attn{4.1} (0.1) &   5.4 (0.1) \\
\bottomrule
\end{tabular}
\end{table}

\section{Discussion} 
We summarize our practical takeaways, in an effort towards creating guidelines
for answering the question: \emph{for a given problem setting, what is the best
  way to produce prediction sets that have good class-conditional coverage but  
  are not too large to be useful?}    

\begin{itemize}
\item \emph{Extremely low-data regime}. When most classes have very few
  calibration examples (say, less than 10), this is not enough data to run
  $\clustered$ or $\classwise$ unless $\alpha$ is large, so the only option is
  to run $\standard$. With this method, $\softmax$ and $\RAPS$ are both good
  score functions. $\RAPS$ tends to yield better class-conditional coverage,
  while $\softmax$ tends to have smaller sets.   

\item \emph{Low-data regime}. When the average number of examples per class is
  low but not tiny (say, around 20 to 75), $\clustered$ tends to strike a good
  balance between variance and granularity and often achieves good
  class-conditional coverage and reasonably-sized prediction sets with either
  $\softmax$ or $\RAPS$. The $\standard$ method with $\RAPS$ is also competitive
  towards the lower-data end of this regime.  

\item \emph{High-data regime.}  When the average number of examples per class is 
  large (say, over 75), $\clustered$ conformal with either $\softmax$ or
  $\RAPS$ continues to do well, and $\classwise$ with these same scores can also
  do well if the classes are balanced. In settings with high class imbalance,
  $\classwise$ is unstable for rare classes and produces excessively large
  prediction sets, whereas $\clustered$ is more robust due to the data-sharing
  it employs.  

\item \emph{Extremely high-data regime}. When the calibration dataset is so
  large that the rarest class has at least, say, 100 examples, then $\classwise$
  with $\softmax$ or $\RAPS$ will be a good choice, regardless of any class
  imbalance.     
\end{itemize}

The boundaries between these regimes are not universal and are dependent on the
data characteristics, and the above guidelines are based only on our findings from
our experiments. We also note that the precise boundary between the low-data and 
high-data regimes is dependent on the particular score function that is used:
the transition happens around 20-40 examples per class for $\softmax$ and 50-100  
for $\RAPS$. This serves as further motivation for $\clustered$, which performs
relatively well in all regimes.  

As a possible direction for future work, it might be interesting to generalize
our approach to the broader problem of group-conditional coverage with many 
groups. In the present setting, the groups are defined by class labels, but our 
clustering methodology could also be applied to other group structures (e.g.,
defined by the input features or components of a mixture distribution). 

\begin{ack}

We thank Margaux Zaffran and Long Nguyen for helpful suggestions that 
improved our paper. 
This work was supported by the National Science Foundation (NSF) Graduate
Research Fellowship Program under grant no.\ 2146752, the European Research
Council (ERC) Synergy Grant Program, the Office of Naval Research (ONR) the
Mathematical Data Science Program, and the ONR Multi-University Research
Initiative (MURI) Program under grant no.\ N00014-20-1-2787.


\end{ack}

\bibliographystyle{plainnat}
\bibliography{references}  

\newpage
\appendix

\section{Proofs} 
\label{sec:proofs}

\paragraph{Proof of Proposition \ref{prop:cluster-conditional-coverage}.}  

For each $m \in \{1,\dots,M\}$, denote by \smash{$G^{(m)}$} the distribution of
the score $s(X,Y)$ conditioned on $Y$ being in cluster $m$. Consider a randomly
sampled test example $(\Xtest, \Ytest)$ with a label in cluster $m$; the test
score \smash{$s_{\mathrm{test}} = s(\Xtest, \Ytest)$} then follows distribution
\smash{$G^{(m)}$}. Next consider \smash{$\{s_i\}_{i \in \cI_2(m)}$}, the scores
for examples in the proper calibration set with labels in cluster $m$; these
also follow distribution \smash{$G^{(m)}$}. Furthermore,
\smash{$s_{\mathrm{test}}$} and the elements of \smash{$\{s_i\}_{i \in 
  \cI_2(m)}$} are all mutually independent, so the result follows by
the standard coverage guarantee for conformal prediction with exchangeable
scores (e.g., see \citet{vovk2005algorithmic} or \citet{lei2018distribution}). 

\paragraph{Proof of Proposition \ref{prop:class-conditional-coverage}.}

This is a direct result of exchangeability and Proposition
\ref{prop:cluster-conditional-coverage}.  

\paragraph{Proof of Proposition \ref{prop:approx-class-conditional-coverage}.}  

Let $S = s(X,Y)$ denote the score of a randomly sampled example $(X,Y) \sim
F$. Fix an $m \in \{1,\dots,M\}$. Define \smash{$\cY^{(m)} = \{y \in \cY :
  \hhat(y) = m\}$} as the set of classes that \smash{$\hhat$} assigns to cluster
$m$. Without a loss of generality, we treat both \smash{$\hhat$} and 
\smash{$\qhat(m)$} as fixed for the remainder of this proof. This can be done by
conditioning on both the clustering and proper calibration sets, leaving only
the randomness in the test point $(X,Y) \sim F$, and then integrating over the
clustering and proper calibration sets in the end.

Let \smash{$G^{(m)}$} denote the distribution of $S$ conditional on \smash{$Y
  \in \cY^{(m)}$}, and let \smash{$S^{(m)} \sim G^{(m)}$}. Similarly, let
\smash{$G^y$} denote the distribution of $S$ conditional on $Y=y$, and let
\smash{$S^y \sim G^y$}. Since we assume that the KS distance between the score
distribution for every pair of classes in cluster $m$ is bounded by $\epsilon$,
and \smash{$G^{(m)}$} is a mixture of these distributions (that is,
\smash{$G^{(m)} = \sum_{y \in \cY^{(m)}} \, \pi_y \cdot G^y$} for some fixed
probability weights $\pi_y$, \smash{$y \in \cY^{(m)}$}), it follows by the 
triangle inequality that        
\[
D_{\mathrm{KS}}(S^y, S^{(m)}) \leq \epsilon, \quad \text{for all $y \in \cY^{(m)}$}.
\]
By definition of KS distance, this implies
\[
\big| \P(S \leq \qhat(m) \mid Y=y) - \P(S \leq \qhat(m) \mid Y \in \cY^{(m)})
\big| \leq \epsilon.
\]
Since the $\clustered$ procedure includes the true label $Y$ from the prediction
set $C$ when \smash{$S \leq \qhat(m)$}, these probabilities can be rewritten
in terms of coverage events:  
\[
\big| \P(Y \in \cC(X) \mid Y=y) - \P(Y \in \cC(X) \mid Y \in \cY^{(m)}) \big|
\leq \epsilon.  
\]
Combining the result from Proposition \ref{prop:cluster-conditional-coverage}
gives the desired conclusion.

\section{Experiment details} 
\label{sec:experiment_details}

\subsection{Score functions} 
\label{sec:score_functions}

We perform experiments using three score functions: 
\begin{itemize}
\item $\softmax$: The softmax-based conformal score at an input $x$ and a label
  $y$ is defined as 
  \[
  s_{\softmax}(x,y) = 1 - \fhat_y(x),
  \] 
  where \smash{$\fhat_y(x)$} is entry $y$ of the softmax vector output by the
  classifier \smash{$\fhat$} at input $x$.

\item $\APS$: The \emph{Adaptive Prediction Sets} (APS) score of
  \citet{romano2020classification} is designed to improve $X$-conditional
  coverage as compared to the more traditional softmax score. This score is
  computed at an input $x$ and label $y$ as follows. Let   
  \[
  \fhat_{(1)}(x) \leq \fhat_{(2)}(x) \leq \cdots \leq \fhat_{(|\cY|)}(x)
  \] 
  denote the sorted values of the base classifier softmax outputs
  \smash{$\fhat_y(x)$}, $y \in \cY$. Let \smash{$k_x(y)$} be the index in
  the sorted order that corresponds to class $y$, that is, 
  \smash{$\fhat_{(k_x(y))} = \fhat_{y}(x)$}.  The APS score is then defined as 
    \[
      s_{\APS}(x,y) = \sum_{i=1}^{k_x(y)-1} \fhat_{(i)}(x) + \mathrm{Unif}([0, 
      \fhat_{(k_x(y))}(x)]). 
    \]

    \item $\RAPS$: The \emph{regularized APS} (RAPS) score of
      \citet{angelopoulos2021uncertainty} is a modification of the APS score
      that adds a regularization term designed to reduce the prediction set
      sizes (which can often be very large with APS). The RAPS score is defined as 
    \[
      s_{\RAPS}(x,y) = s_{\mathsf{APS}}(x,y) + \max(0, \lambda (k_x(y) -
      k_{\mathrm{reg}})),
    \]
    where \smash{$k_x(y)$} is as defined above, and $\lambda$ and
    $k_{\mathrm{reg}}$ are user-chosen parameters. In all of our experiments, we 
    use $\lambda = 0.01$ and $k_{\mathrm{reg}} = 5$, which
    \citet{angelopoulos2021uncertainty} found to work well for ImageNet. 
\end{itemize}

\subsection{Model training} 
\label{sec:model_training}

An important consideration when we fine-tune and calibrate our models is that we
must reserve sufficient data to evaluate the class-conditional coverage of
the conformal methods. This means we aim to exclude at least 250 examples per
class from the fine-tuning and calibration sets so that we can then use this
data for validation (applying the conformal methods and computing coverage and
set size metrics).  

For all data sets except ImageNet, we use a ResNet-50 model as our base
classifier. We initialize to the \verb|IMAGENET1K_V2| pre-trained weights from
\verb|PyTorch| \citep{paszke2019pytorch}, and then fine-tune all parameters by 
training on the data set at hand. For ImageNet, we must do something different,
as explained below.

\paragraph{ImageNet.} 

Setting up this data set for our experiments is a bit tricky because we need 
sufficient data for performing validation, but we also need this data to be
separate from the fine-tuning and calibration sets. The ImageNet validation set
only contains 50 examples per class, which is not enough for our setting. The
ImageNet training set is much larger, with roughly 1000 examples per class, but
if we want to use this data for validation, then we cannot use the ResNet-50
initialized to the \verb|IMAGENET1K_V2| pre-trained weights, as these weights
were obtained by training on the whole ImageNet training set. We
therefore instead use a SimCLR-v2 model \citep{chen2020big}, which was trained
on the ImageNet training set \emph{without labels}, to extract feature vectors
of length 6144 for all images in the ImageNet training set. We then use 10\% of
these feature vectors for training a linear head (a single fully connected
neural network layer). After training for 10 epochs, the model achieves a
validation accuracy of 78\%. We then apply the linear head to the remaining 90\%
of the feature vectors to obtain softmax scores for the calibration set.

\paragraph{CIFAR-100.} 

This data set has 600 images per class (500 from the training set and 100 from 
the validation set). We combine the data and then randomly sample 50\% for
fine-tuning, and we use the remaining data for calibrating and validating our 
procedures. After training for 30 epochs, the validation accuracy is 60\%.  

\paragraph{Places365.} 

This data set contains more than 10 million images of 365 classes, where each
class has 5000 to 30000 examples. We randomly sample 90\% of the data for
fine-tuning, and we use the remaining data for calibrating and validating our
procedures. After training for one epoch, the validation accuracy is 52\%.

\paragraph{iNaturalist.} 

This data set has class labels of varying specificity. For example, at the
\verb|species| level, there are 6414 classes with 300 examples each (290
training examples and 10 validation examples) and a total of 10000 classes with
at least 150 examples. We instead work at the \verb|family| level, which groups
the species into 1103 classes. We randomly sample 50\% of the data for
fine-tuning, and we use the remaining for calibrating and validating our
procedures. After training for one epoch, the validation accuracy is 69\%.    

However, due to high class imbalance and sampling randomness, some classes have
insufficient validation samples, so we filter out classes with fewer than 250
validation examples, which leaves us with 633 classes. The entries of the
softmax vectors that correspond to rare classes are removed and the vector is
renormalized to sum to one.  


\subsection{Measuring class balance in Table \ref{tab:dataset_summary}}
\label{sec:class_balance_metric}

The class balance metric in Table \ref{tab:dataset_summary} is defined as the
number of examples in the rarest 5\% of classes divided by the expected number
of examples if the class distribution were perfectly uniform. This metric is
bounded between 0 and 1, with lower values denoting more class imbalance. 
We compute this metric using \smash{$D_{\mathrm{fine}}^c$}.

\subsection{Choosing clustering parameters} 
\label{sec:clustering_parameters}

To choose $\gamma \in [0,1]$ and $M \geq 1$ for $\clustered$, as described
briefly in Section \ref{sec:experimental_setup}, we employ two intuitive
heuristics. We restate these heuristics in more detail here.   

\begin{itemize}
\item First, to distinguish between more clusters (or distributions), we need
  more samples from each distribution. As a rough guess, to distinguish between
  two distributions, we want at least four samples per distribution; to
  distinguish between five distributions, we want at least ten samples per
  distribution. In other words, we want the number of clustering examples per
  class to be at least twice the number of clusters. This heuristic can be
  expressed as  
  \begin{equation} 
  \label{eq:heuristic1}
  \gamma \tilde{n} \geq 2M,
  \end{equation}
  where \smash{$\gamma \tilde{n}$} is the expected number of clustering examples
  for the rarest class that is not assigned to the null cluster.  

\item Second, we want enough data for computing the conformal quantiles for each
  cluster. Specifically, we seek at least 150 examples per cluster on
  average. This heuristic can be expressed as    
  \begin{equation} 
  \label{eq:heuristic2}
  (1-\gamma) \tilde{n} \frac{K}{M} \geq 150,
  \end{equation}
  where $K/M$ is the average number of classes per cluster and
  \smash{$(1-\gamma) \tilde{n}$} is the expected number of proper calibration
  examples for the rarest class not assigned to the null cluster.  
\end{itemize}

\noindent
Changing the inequalities of \eqref{eq:heuristic1} and \eqref{eq:heuristic2}
into equalities and solving for $\gamma$ and $M$ yields 
\[
M = \frac{\gamma \tilde{n}}{2} \qquad \text{and} \qquad \gamma =
\frac{K}{K+75}. 
\]

\paragraph{Varying the clustering parameters.} 

\begin{figure}[htb]
\centering
\includegraphics[width=\textwidth]{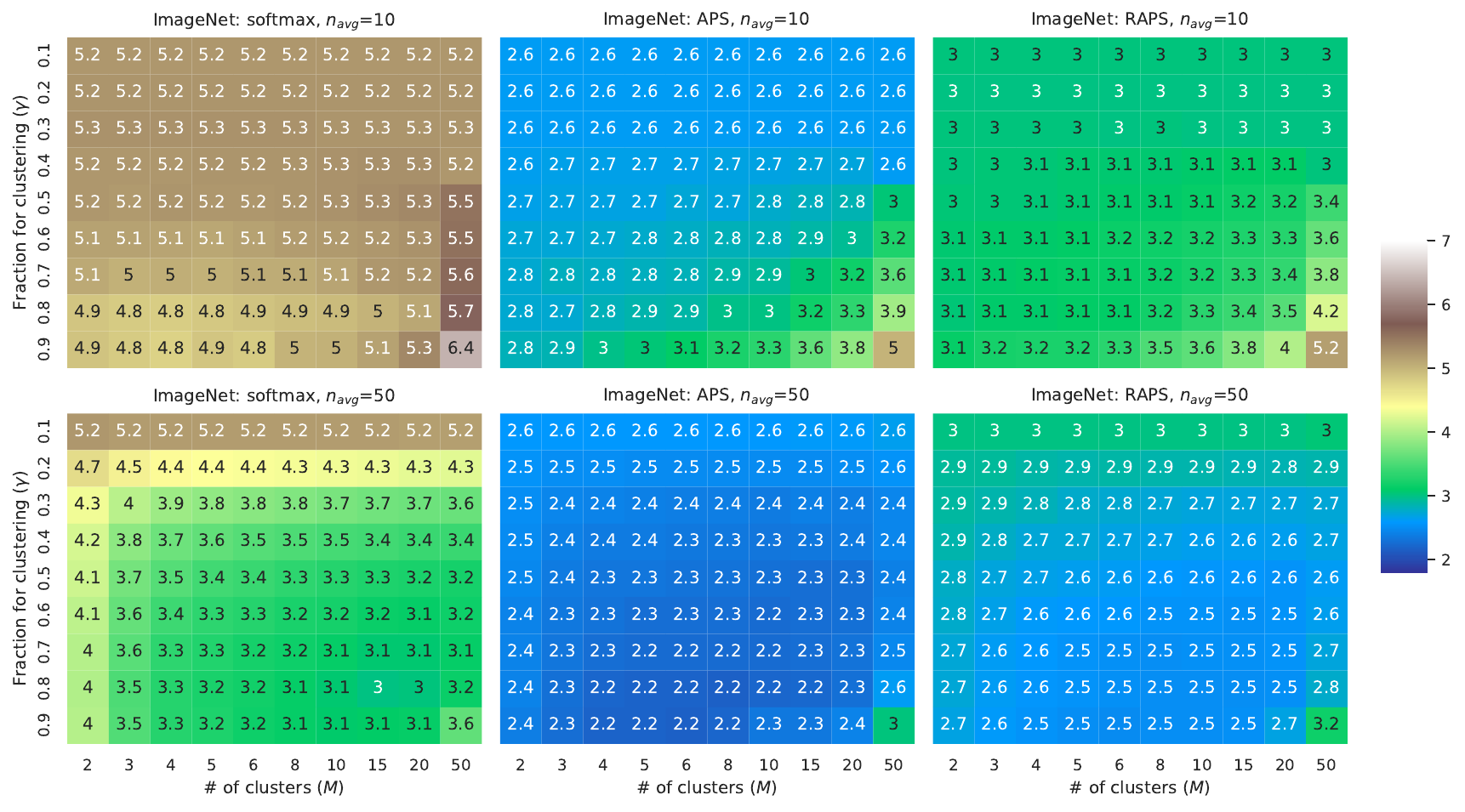}
\caption{Average class coverage gap on ImageNet for $\navg \in \{10, 50\}$,
  using the $\softmax$, $\APS$, and $\RAPS$ scores, as we vary the clustering 
  parameters. Each entry is averaged across 10 random splits of the data into
  calibration and validation sets.}  
\label{fig:heatmaps}
\end{figure}

As sensitivity analysis, we examine the performance of $\clustered$ across a
wide range of values for the tuning parameters $\gamma$ and $M$. As the heatmaps
in Figure \ref{fig:heatmaps} confirm, the performance of $\clustered$ is not
particularly sensitive to the values of these parameters. When $\navg=10$, the
heuristic chooses $\gamma=0.89$ and $M=4$, meanwhile, when $\navg=50$, the
heuristic chooses $\gamma \in [0.88, 0.92]$ and $M \in [7,12]$ (since the
calibration data set is randomly sampled, and $\gamma$ and $M$ are chosen based
on the calibration data set, there is randomness in the chosen values).
However, there are large areas surrounding these values that would yield similar
performance. We observe that the heuristics do not always choose the parameter
values that yield the lowest CovGap. The heatmaps reveal that the optimal
parameter values are dependent not only on data set characteristics, but also on
the score function. Future work could be done to extract further performance
improvements by determining a better method for choosing $\gamma$ and $M$.

\section{Additional experimental results}

We present additional experimental results in this section. As in the main text,
the shaded regions in plots denote $\pm 1.96$ times the standard errors. 

\subsection{$\RAPS$ CovGap results} 
\label{sec:results_other_score_functions}

\begin{figure}[htb]
\centering
\includegraphics[width=0.965\textwidth]{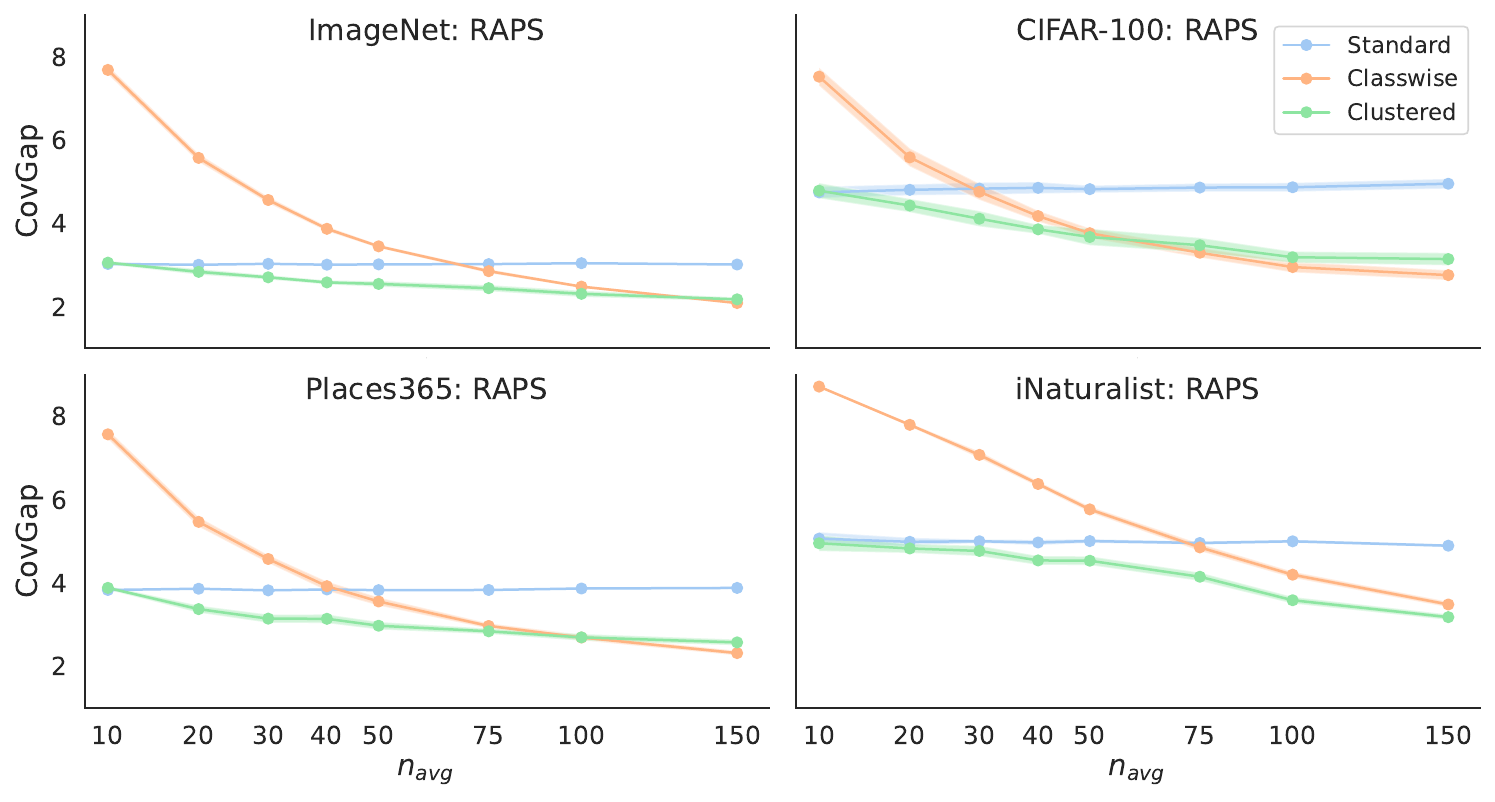} 
\caption{Average class coverage gap for ImageNet, CIFAR-100, Places365, and
  iNaturalist, using the $\RAPS$ score, as we vary the average number of
  calibration examples per class.}
\label{fig:RAPS_alldatasets_covgap}
\end{figure}

Figure \ref{fig:RAPS_alldatasets_covgap} shows the CovGap on all data sets when
we use $\RAPS$ as our score function.  

\subsection{Additional metrics} 
\label{sec:additional_metrics}

\paragraph{Average set size.} 

\begin{figure}[htbp]
\centering
\includegraphics[width=0.965\textwidth]{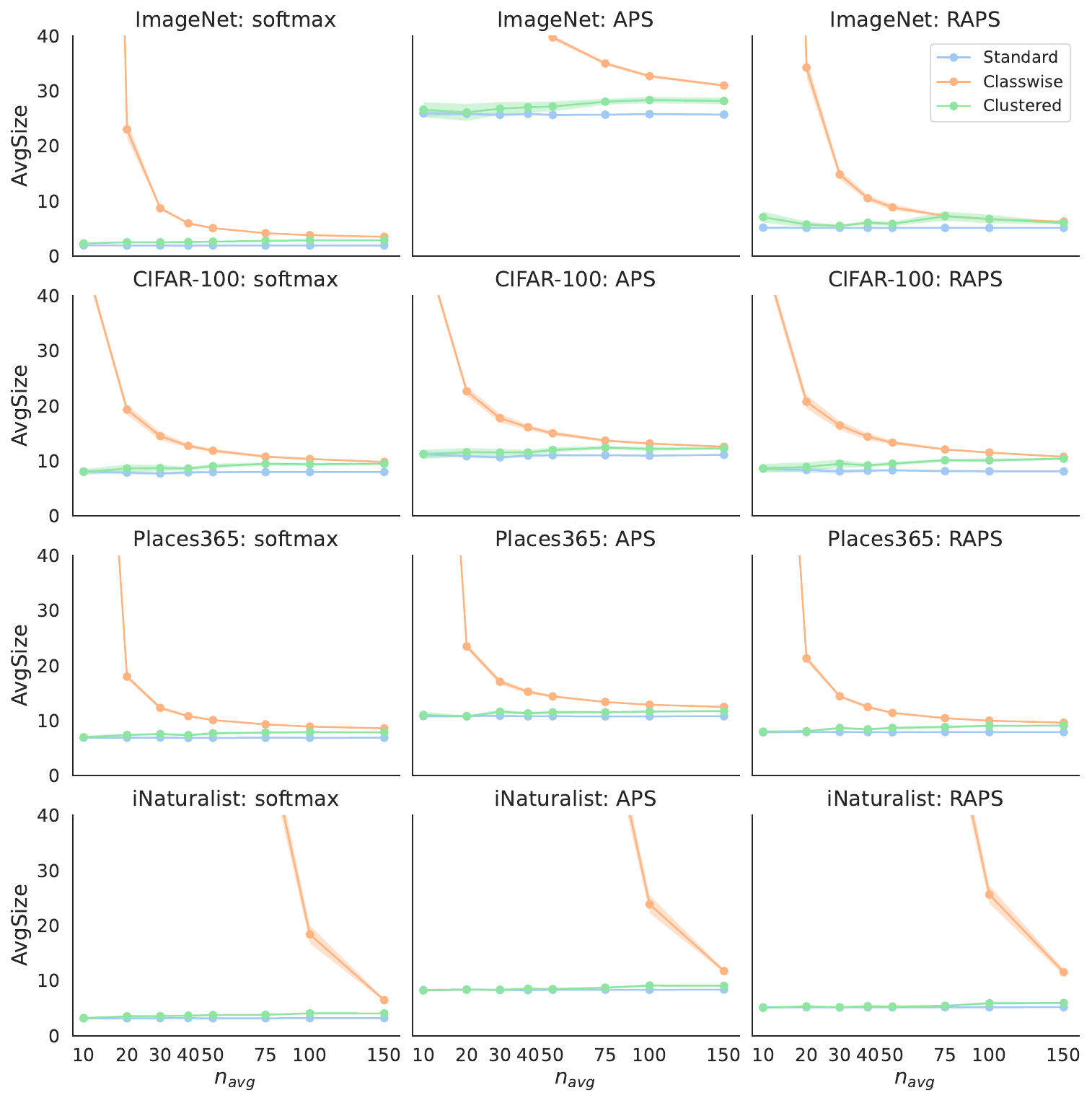} 
\caption{Average set size for ImageNet, CIFAR-100, Places365, and iNaturalist,
  for the $\softmax$, $\APS$, and $\RAPS$ scores, as we vary the average number 
  of calibration examples per class.}   
\label{fig:avg_set_size}
\end{figure}

To supplement Table \ref{tab:results} from the main text, which reports AvgSize
for select values of $\navg$, Figure \ref{fig:avg_set_size} plots AvgSize for all
values of $\navg$ that we use in our experimental setup. Note that $\RAPS$
sharply reduces AvgSize relative to $\APS$ on ImageNet and also induces a slight
reduction for the other three data sets. This asymmetric reduction is likely due
to the fact that the $\RAPS$ hyperparameters, which control the strength of the
set size regularization, were tuned on ImageNet. The set sizes of $\RAPS$ on
other data sets could likely be improved by tuning the hyperparameters for each
data set.    

\paragraph{Fraction undercovered.} 

\begin{figure}[htbp]
\centering
\includegraphics[width=0.965\textwidth]{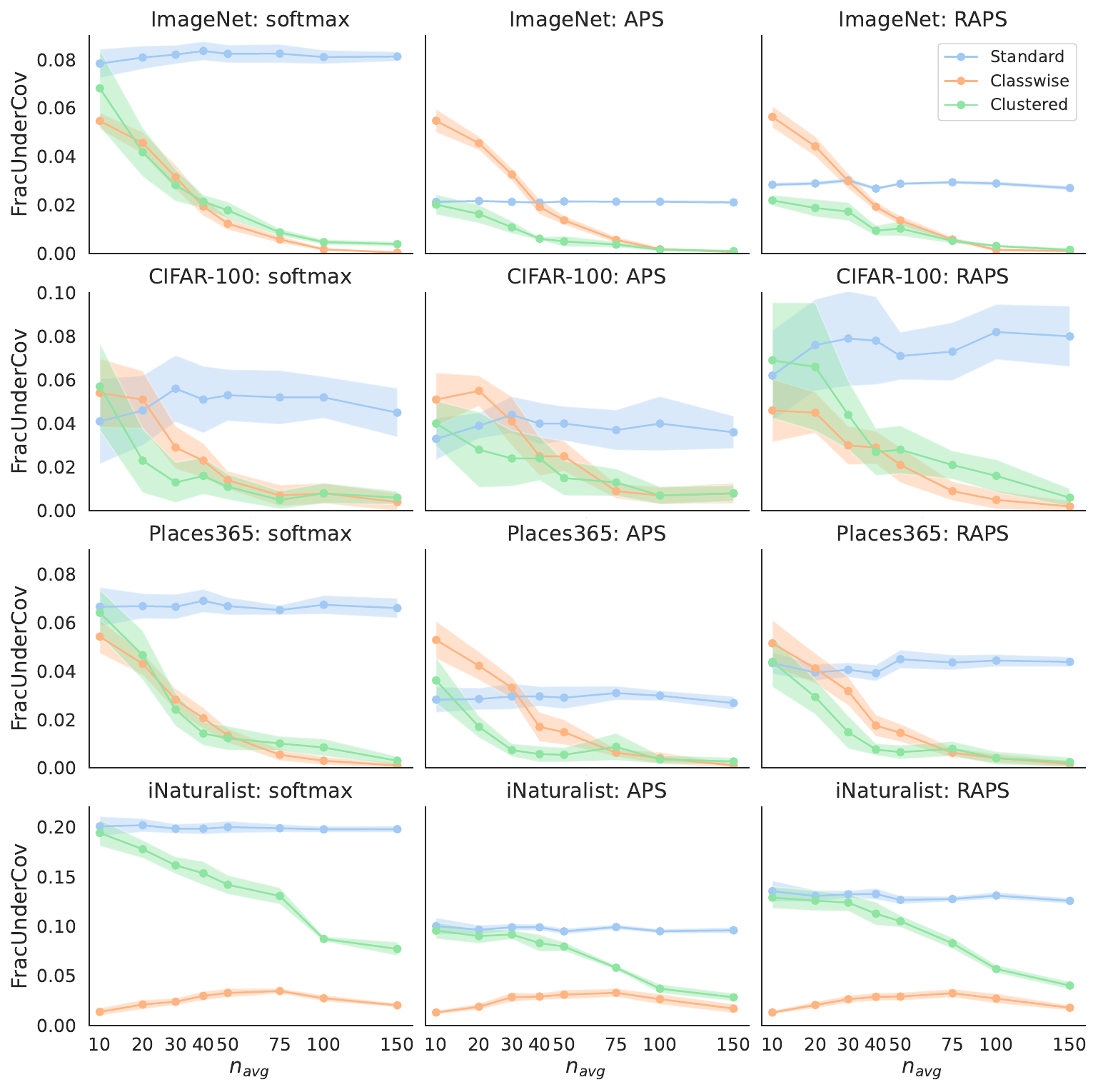} 
\caption{Fraction of severely undercovered classes for ImageNet, CIFAR-100,
  Places365, and iNaturalist, using the $\softmax$, $\APS$, and $\RAPS$
  scores, as we vary the average number of calibration examples per class.}   
\label{fig:very_undercovered}
\end{figure}

In various practical applications, we want to limit the number of classes that
are severely undercovered, which we define as having a class-conditional
coverage more than 10\% below the desired coverage level. We define the fraction
of undercovered classes (FracUnderCov) metric as:    
\[
\text{FracUnderCov} = \frac{1}{|\cY|}\sum_{y=1}^{|\cY|} \ind{\hat{c}_y \leq 1-\alpha-0.1},
\]
recalling that \smash{$\hat{c}_y$} is the empirical class-conditional coverage
for class $y$. Figure \ref{fig:very_undercovered} plots FracUnderCov for all
experimental settings. Comparing to the CovGap plots in Figure
\ref{fig:alldatasets_covgap} and Figure \ref{fig:RAPS_alldatasets_covgap}, we
see that the trends in FracUnderCov generally mirror the trends in
CovGap. However, FracUnderCov is a much noisier metric, as evidenced by the
large error bars. Another difference is that FracUnderCov as a metric is unable
to penalize uninformatively large set sizes. This is best seen in the
performance of $\classwise$ on iNaturalist: for every score function,
$\classwise$ has very low FracUnderCov, but this is achieved by producing
extremely large prediction sets, as shown in the bottom row of Figure
\ref{fig:avg_set_size}. Meanwhile, CovGap is somewhat able to penalize this kind
of behavior since unnecessarily large set sizes often lead to overcoverage, and
CovGap penalizes overcoverage.    

\subsection{Auxiliary randomization} 
\label{sec:randomized_methods}

The conformal methods in the main paper generate \emph{deterministic} prediction
sets, so running the method on the same input will always produce the same
prediction set. These prediction sets are designed to achieve \emph{at least}
$1-\alpha$ marginal, class-conditional, or cluster-conditional coverage. In most
practical situations, it is arguably undesirable to use non-deterministic or
\emph{randomized} prediction sets (say, if you are a patient, you would not
want your doctor to tell you that your diagnoses change depending on a random
seed). However, if one is willing to accept randomized prediction sets, then the
conformal methods described in the main text can be modified to achieve
\emph{exact} $1-\alpha$ coverage.   

\paragraph{Randomizing to achieve exact $1-\alpha$ coverage.} 

Recall that the unrandomized $\standard$ method used in the main paper uses 
\[
\qhat = \quantile\bigg( \frac{\lceil(N+1)(1-\alpha)\rceil}{N}, \{s_i\}_{i=1}^N
\bigg), 
\]
which yields a coverage guarantee of  
\[
\P(\Ytest \in \cC(\Xtest) \mid \Ytest = y) = \frac{\lceil (N+1)(1-\alpha)
  \rceil}{N+1} \geq 1-\alpha, 
\]
under the assumption that the scores are distinct almost surely. The equality
above holds because the event that $\Ytest$ is included in $\cC(\Xtest)$ is
equivalent to the event that $s(\Xtest, \Ytest)$ is one of the $\lceil
(N+1)(1-\alpha) \rceil$ smallest scores in the set containing the calibration
points and the test point, and, by exchangeability, this occurs with probability
exactly $\lceil (N+1)(1-\alpha) \rceil / (N+1)$. By similar reasoning, if we
instead use  
\begin{equation} 
\label{eq:qtilde}
\qtilde = \quantile\bigg( \frac{\lceil(N+1)(1-\alpha) \rceil-1}{N},
\{s_i\}_{i=1}^N \bigg ) 
\end{equation}
as our conformal quantile (note the added $-1$ in the numerator), then we would have
\[
\P(\Ytest \in \cC(\Xtest) \mid \Ytest = y) = \frac{\lceil (N+1)(1-\alpha) 
  \rceil-1}{N+1} < 1-\alpha.   
\]
To summarize, using \smash{$\qhat$} results in at least $1-\alpha$ coverage, and
using \smash{$\qtilde$} results in less than $1-\alpha$ coverage.  Thus, to
achieve exact $1-\alpha$ coverage, we can randomize between using
\smash{$\qhat$} and \smash{$\qtilde$}. Let
\[
b = \frac{\lceil (N+1)(1-\alpha) \rceil}{N+1} - (1-\alpha)
\]
be the amount by which the coverage using \smash{$\qhat$} overshoots the desired
coverage level and let  
\[
c = (1-\alpha) -\frac{\lceil (N+1)(1-\alpha)\rceil-1}{N+1} 
\]
be the amount by which the coverage using \smash{$\qtilde$} undershoots the
desired coverage level. Then, if we define the Bernoulli random variable 
\[
B \sim \mathrm{Bern}\bigg( \frac{c}{b+c} \bigg),
\]
independent of everything else that is random, and set 
\[
\qhat_{\mathrm{rand}} = B\qhat + (1-B) \qtilde
\]
then the prediction sets created using \smash{$\qhat_{\mathrm{rand}}$} will have
exact $1-\alpha$ marginal coverage.
The same idea translates to $\classwise$ and $\clustered$ methods (where we
randomize \smash{$\qhat^y$} and \smash{$\qhat(m)$}, respectively).   

\begin{figure}[htbp]
\centering
\includegraphics[width=0.965\textwidth]{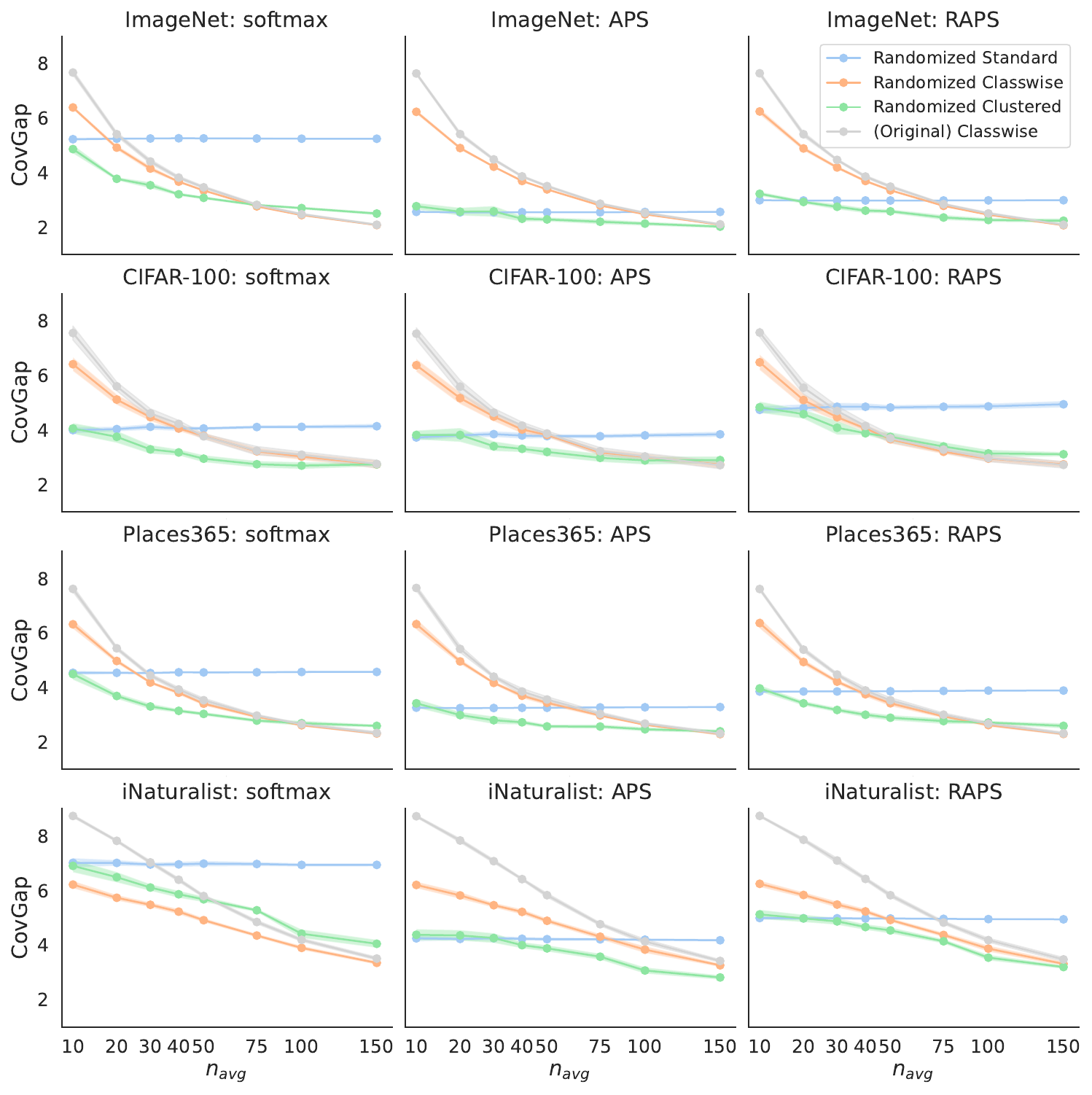}  
\caption{Average class coverage gap for randomized methods on ImageNet,
  CIFAR-100, Places365, and iNaturalist, for the $\softmax$, $\APS$, and $\RAPS$
  scores, as we vary the average number of calibration examples per class. The
  unrandomized original $\classwise$ method is also plotted for comparison
  purposes.}   
\label{fig:randomized_covgap}
\end{figure}
\begin{figure}[htbp]
\centering
\includegraphics[width=0.965\textwidth]{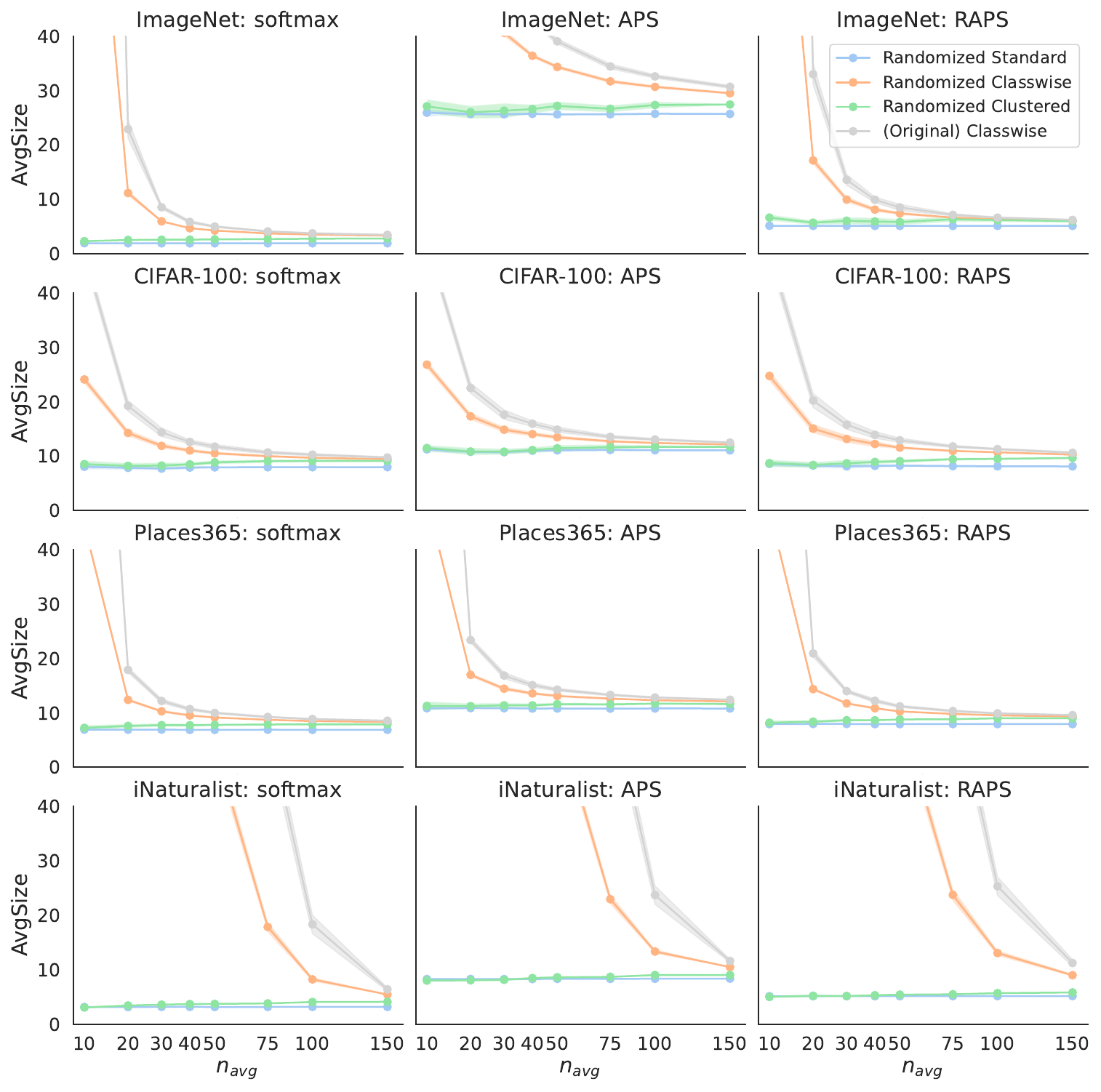} 
\caption{Average set size for randomized methods on ImageNet, CIFAR-100,
  Places365, and iNaturalist, for the $\softmax$, $\APS$, and $\RAPS$ scores, as
  we vary the average number of calibration examples per class. The unrandomized
  original $\classwise$ method is also plotted for comparison purposes.}  
\label{fig:randomized_avg_set_size}
\end{figure}
\begin{figure}[htbp]
\centering
\includegraphics[width=0.965\textwidth]{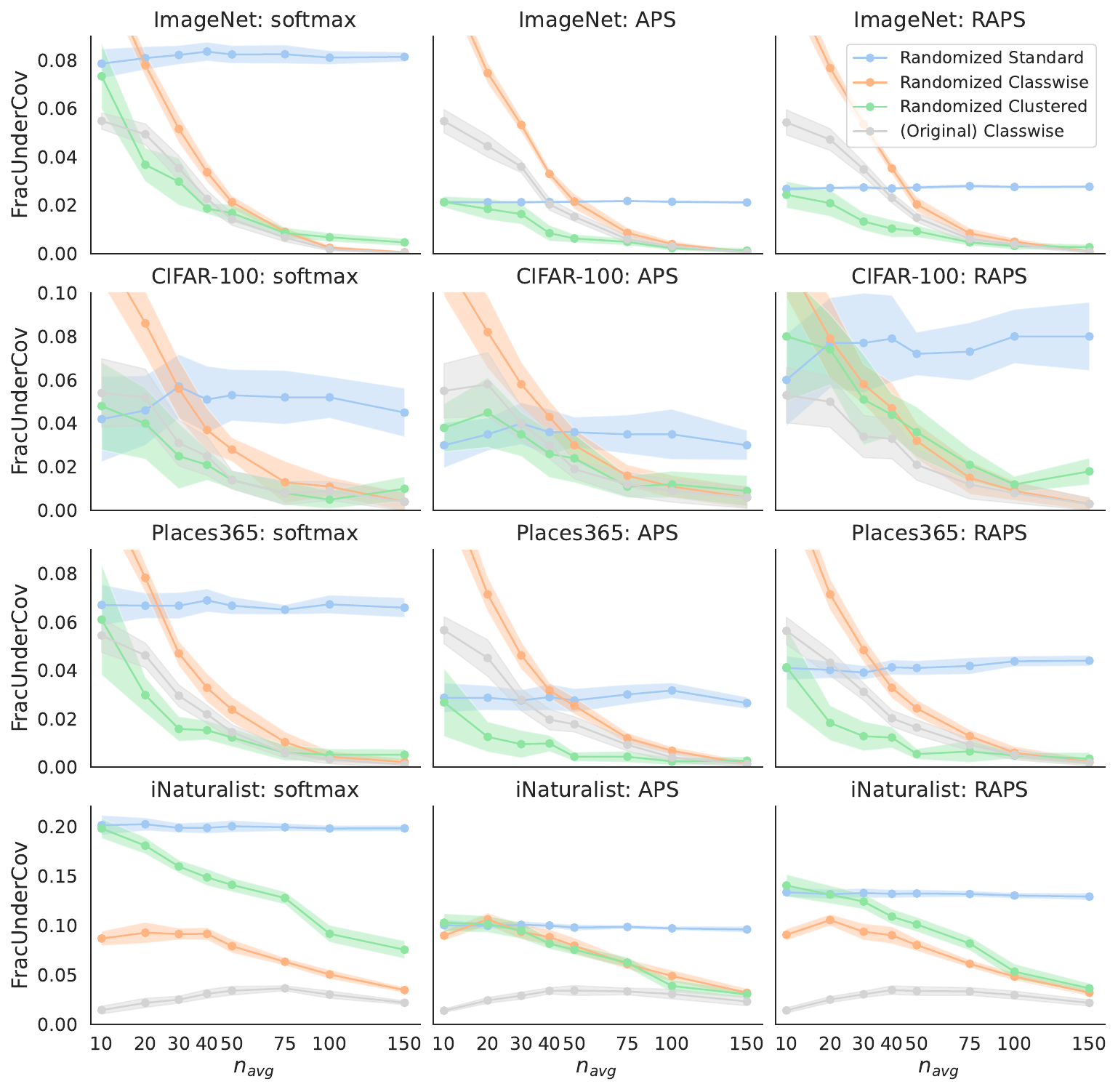} 
\caption{Fraction of severely undercovered classes for randomized methods on
  ImageNet, CIFAR-100, Pla- ces365, and iNaturalist, for the $\softmax$, $\APS$, 
  and $\RAPS$ scores, as we vary the average number of calibration examples per
  class. The unrandomized original $\classwise$ method is also plotted for
  comparison purposes.}    
\label{fig:randomized_very_undercovered}
\end{figure}

Figures \ref{fig:randomized_covgap}, \ref{fig:randomized_avg_set_size}, and
\ref{fig:randomized_very_undercovered} display the CovGap, AvgSize, and
FracUnderCov for the randomized versions of the conformal methods. Comparing
against earlier plots, we observe that the performance of unrandomized and
randomized   $\standard$ and $\clustered$ are essentially identical in terms of
all three metrics. However, we cam see that randomized $\classwise$ exhibits a
large improvement relative to unrandomized $\classwise$ in terms of CovGap and 
AvgSize. That said, the previously-given qualitative conclusions do not change,
and the set sizes of randomized $\classwise$ are still too large to be
practically useful.  

\end{document}